\documentclass[10pt,twocolumn,letterpaper]{article}

\usepackage[pagenumbers,final]{cvpr}

\newcommand{\todob}[2][]{}
\newcommand{\todobin}[2][]{}
\newcommand{\todonikos}[2][]{}
\newcommand{\todosm}[2][]{}
\newcommand{\todoys}[2][]{}
\newcommand{\todoks}[2][]{}

\newcommand{\TODO}[1]{\textbf{\color{red}[TODO: #1]}}
\renewcommand{\TODO}[1]{}

\usepackage{microtype}

\usepackage{xspace}

\usepackage{soul}
\setuldepth{foobar}

\usepackage[utf8]{inputenc}
\usepackage[T1]{fontenc}    
\usepackage[english]{babel}
\usepackage{tcolorbox}
\tcbuselibrary{skins}
\usepackage{listings}
\usepackage{wrapfig}
\usepackage{algorithm}
\usepackage{algpseudocodex}
\usepackage{array}
\usepackage{booktabs}
\usepackage{dblfloatfix}
\usepackage{placeins}

\DeclareMathOperator*{\mean}{mean}
\DeclareMathOperator*{\std}{std}
\DeclareMathOperator{\clip}{clip}
\DeclareMathOperator{\SNR}{SNR}

\DeclareMathOperator{\Var}{Var}
\newcommand{\KL}{D_{\mathrm{KL}}}
\newcommand{\tref}{\text{ref}}

\renewcommand\bar\overline

\DeclareMathOperator{\E}{\mathbb{E}}
\DeclareMathOperator{\var}{Var}

\newcommand{\calL}{\ensuremath{\mathcal{L}}}

\newcommand{\calN}{\ensuremath{\mathcal{N}}}

\def\nd/{\textsuperscript{nd}}
\def\rd/{\textsuperscript{rd}}
\def\th/{\textsuperscript{th}}

\newcommand{\eps}{\ensuremath{\epsilon}}

\makeatletter
\def\nnil{\nil}
\newcounter{prob}

\newcounter{dual}

\makeatother

\newenvironment{prob*}{%
	\csname equation*\endcsname%
	\aligned%
}{%
	\endaligned%
	\csname endequation*\endcsname%
}

\definecolor{cvprblue}{rgb}{0.21,0.49,0.74}
\usepackage[pagebackref,breaklinks,colorlinks,allcolors=cvprblue]{hyperref}
\newcommand{\kr}[1]{}
\newcommand{\ys}[1]{}
\def\confName{CVPR}
\def\confYear{2026}

\title{Stepwise Credit Assignment for GRPO on Flow-Matching Models}

\author{%
Yash Savani\textsuperscript{\dag\,*}\quad Branislav Kveton\textsuperscript{\ddag}\quad Yuchen Liu\textsuperscript{\ddag}\quad Yilin Wang\textsuperscript{\ddag}\\
Jing Shi\textsuperscript{\ddag}\quad Subhojyoti Mukherjee\textsuperscript{\ddag}\quad Nikos Vlassis\textsuperscript{\ddag}\quad Krishna Kumar Singh\textsuperscript{\ddag}\\
\textsuperscript{\dag}Carnegie Mellon University\quad \textsuperscript{\ddag}Adobe Research\\[2pt]
{\small \textsuperscript{*}Work done while intern at Adobe. Corresponding author: \texttt{ysavani@cs.cmu.edu}.}\\ 
{\small \url{https://stepwiseflowgrpo.com}}
}

\begin{document}

\twocolumn[{
\renewcommand\twocolumn[1][]{#1}%
\maketitle
\begin{center}
    \centering
    \includegraphics[width=0.80\linewidth]{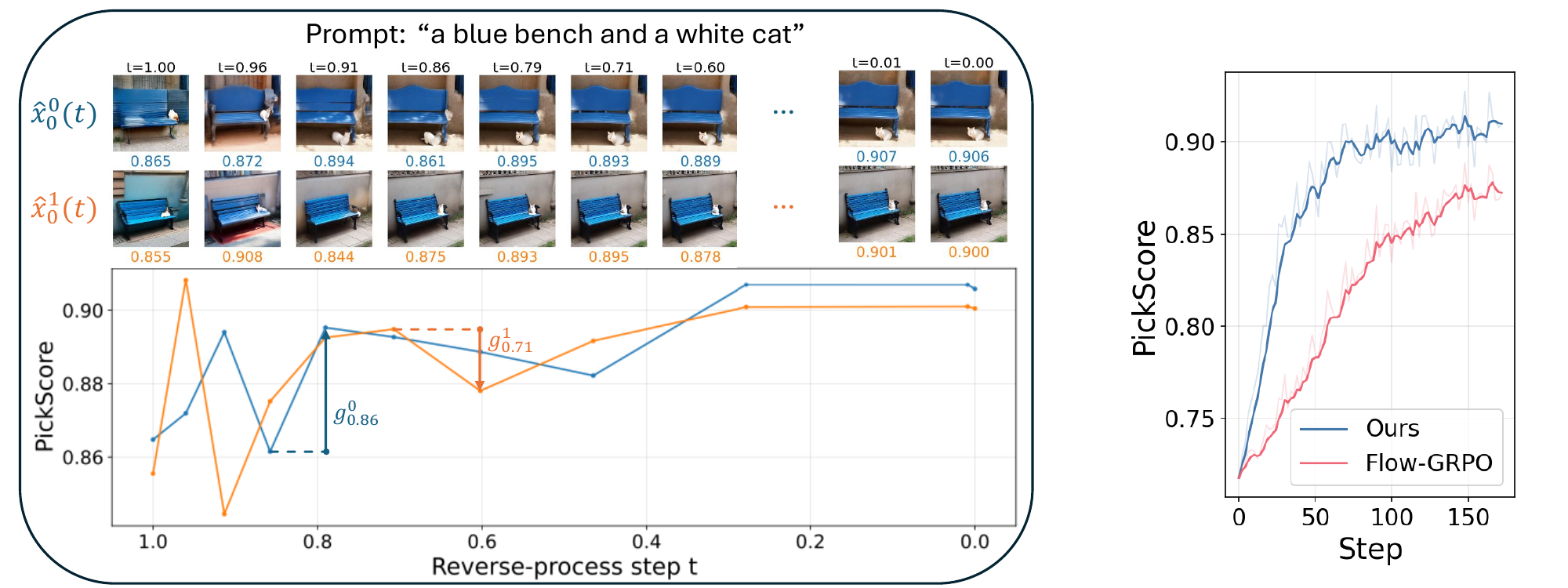}
    \captionof{figure}{\textbf{Stepwise credit assignment from temporal reward structure.}
\textbf{(Left)} Two trajectories for the same prompt, showing Tweedie estimates
$\hat{x}_0^i(t)$ and their PickScore rewards $r_t^i$ at each denoising step.
The reward curves are non-monotonic and frequently cross---trajectory 0 (blue)
dips at $t{=}0.86$ before recovering, while trajectory 1 (orange) drops sharply
at $t{=}0.71$---yet both reach similar final rewards (${\sim}0.90$). Uniform
credit assignment would treat these trajectories nearly identically, reinforcing
the poor intermediate steps along with the good ones. Stepwise-Flow-GRPO instead
uses gains $g_t^i = r_{t-1}^i - r_t^i$ to penalize steps that hurt reward and
credit steps that improve it, regardless of final outcome.
\textbf{(Right)} This finer credit assignment yields faster convergence and
higher final reward compared to Flow-GRPO.}

    \vspace{0pt}
    \label{fig:teaser}
\end{center}
}]

\begin{abstract}
Flow-GRPO successfully applies reinforcement learning to flow models, but uses \emph{uniform credit assignment} across all steps. This ignores the temporal structure of diffusion generation: early steps determine composition and content (low-frequency structure), while late steps resolve details and textures (high-frequency details). Moreover, assigning uniform credit based solely on the final image can inadvertently \emph{reward suboptimal intermediate steps}, especially when errors are corrected later in the diffusion trajectory. We propose \textbf{Stepwise-Flow-GRPO}, which assigns credit based on each step's reward improvement. By leveraging Tweedie's formula to obtain intermediate reward estimates and introducing gain-based advantages, our method achieves superior sample efficiency and faster convergence. We also introduce a DDIM-inspired SDE that improves reward quality while preserving stochasticity for policy gradients.
\end{abstract}

\section{Introduction}

Reinforcement learning (RL) has dramatically improved reasoning in large language models (LLMs) \cite{shao24deepseekmath}, but adapting it to flow-matching models for text-to-image generation remains challenging. \textbf{Flow-GRPO}~\cite{liu2025flow} and \textbf{Dance-GRPO}~\cite{xue2025dancegrpo} showed that policy gradients can be applied to flow-matching models like \texttt{SD-3.5-M} by converting the deterministic flow ordinary differential equation (ODE) into a stochastic differential equation (SDE) with matched marginals. These approaches assign the same advantage---calculated from the rewards on the final images---to all steps in a trajectory, effectively rewarding or penalizing every denoising step based \emph{solely on the final generated image.}

This uniform credit assignment ignores the \emph{temporal structure} of all diffusion generation processes, where different steps contribute to distinct image properties.
Early denoising steps contain primarily low-frequency information, which determines composition and layout, while late steps resolve high-frequency details and textures. Rewarding an entire trajectory based on final image quality conflates these phases, potentially reinforcing poor early decisions that are compensated for later. As shown in~\cref{fig:teaser}, intermediate reward estimates reveal substantial temporal structure across steps that uniform credit assignment fails to exploit.

To capitalize on this opportunity, we propose \textbf{Stepwise-Flow-GRPO}, a stepwise credit assignment approach that directly rewards the contribution of each denoising step. Using Tweedie's formula~\cite{efron2011tweedie}, we predict $\hat{x}_0(t) := \E[x_0|x_t]$ from intermediate noisy states $x_t$ and reward these estimates as $r_t := R(\hat x_0(t), \text{prompt})$. We then use the stepwise reward gain $g_t = r_{t-1} - r_t$---measuring the amount each step improves the reward---to calculate the advantages that are used to optimize the policy with GRPO~\cite{shao24deepseekmath}. This rewards each step based on its actual contribution rather than final image quality alone. This design offers better credit assignment and improves sample efficiency. \cref{fig:flowsde_results} shows that Stepwise-Flow-GRPO achieves better sample efficiency during training and converges faster.

We also notice that while the Flow-GRPO SDE provably matches marginals of the flow ODE, the images generated with the SDE are noisy. To account for this, we introduce a new SDE, inspired by DDIM~\cite{song2021denoising}, that produces high-quality images while matching the Flow-ODE Fokker-Planck marginal in the Taylor limit. Since reward models are trained on clean images, the noisy samples from Flow-GRPO's SDE degrade reward signal; our formulation mitigates this, providing a complementary way to accelerate Flow-GRPO. We make the following contributions:
\begin{itemize}
  \item We introduce Stepwise-Flow-GRPO, which rewards relative gain for each diffusion step instead of just rewarding final image.
  \item We further optimize Flow-GRPO using an improved SDE inspired by DDIM to produce less noisy samples.
  \item Our results show that Stepwise-Flow-GRPO achieves significantly better sample efficiency and convergence rates than standard Flow-GRPO.
\end{itemize}

\section{Related Work}

\noindent \textbf{Reinforcement learning (RL).} RL is an area of machine learning where the goal is to learn a policy that maximizes a long-term reward in an uncertain environment \citep{sutton98reinforcement}. The beginnings of RL can be traced to model-based approaches for acting under uncertainty, such as \emph{Markov decision processes (MDPs)} \citep{bellman57dynamic,puterman94markov} and \emph{partially-observable MDPs (POMDPs)} \citep{sondik71thesis}. Because of the generality of the framework, many RL algorithms have been proposed, such as temporal-difference learning \citep{sutton88learning}, Q-learning \citep{watkins92qlearning}, policy gradients \citep{williams92simple}, and actor-critic methods \citep{sutton00policy}. The main challenge with applying classic RL algorithms, which rely on value functions \cite{bellman57dynamic} and Q functions \citep{watkins92qlearning}, to modern generative AI models is their scale. Therefore, most modern approaches to RL are based on policy gradients with KL regularization \citep{todorov06linearlysolvable,schulman15trust}. In \emph{proximal policy optimization (PPO)} \citep{schulman17proximal}, a KL-regularized policy is optimized with respect to a reward model. This approach has been successfully applied to reinforcement learning from human feedback \citep{ouyang22training}. The main challenge with applying PPO in practice is that it requires a fine-grained critic model, which is yet another challenging learning problem to solve. \emph{Group relative policy optimization (GRPO)} \citep{shao24deepseekmath} overcomes this problem by estimating and standardizing the rewards using simulation against an environment.

Our work replaces the total trajectory reward in GRPO with per-step gains, yielding fine-grained credit assignment analogous to a learned critic in PPO---but without requiring a separate critic model. This design is motivated by \citet{kveton25adaptive}, who showed that KL-regularized policy gradients over stepwise gains can learn near-optimal greedy policies when the reward is monotone and submodular in the state, extending classic guarantees from submodular optimization \citep{nemhauser78approximation,golovin11adaptive}. We are the first to apply this gain-based transformation to GRPO and flow models, generalizing beyond the on-policy setting of \citet{kveton25adaptive} to off-policy optimization with group-relative advantages.

\noindent \textbf{RL for image generation.} Recent progress in diffusion models~\cite{dhariwal2021diffusion,Rombach_2022_CVPR,podell2024sdxl,chen2023pixartalpha,saharia2022photorealistic} has led to substantial improvements in text-to-image generation, outperforming earlier GAN~\cite{goodfellow2014generative,xu2017attngan,kang2023gigagan} and VAE-based~\cite{pmlr-v139-ramesh21a} methods in both image quality and semantic alignment. Among these, flow-matching techniques~\cite{Esser2024SD3,labs2025flux1kontextflowmatching} have shown strong performance but still face challenges in achieving precise image-text alignment and maintaining aesthetic fidelity~\cite{huang2023t2icompbench,wu2024conceptmix,chatterjee2024getting}. To address these issues, researchers have explored methods such as attention manipulation~\cite{chefer2023attendandexcite,Agarwal2023ASTARTA,bao2024sepen}, contrastive learning~\cite{lee2025aligningtextimagediffusion}, and asynchronous diffusion~\cite{hu2025asynchronousdenoisingdiffusionmodels}. While these approaches offer partial improvements, they often lack robustness and generalization.

More recently, RL has emerged as a promising direction for enhancing alignment and image quality~\cite{jiang2025t2i,pan2025focusdiff,xue2025dancegrpo, liu2025flow}, which leverages pre-trained VLM models~\cite{pickscore,imagereward,unifiedreward,ma2025hpsv3widespectrumhumanpreference} to measure various aspects related to image quality and text alignment. Flow-GRPO~\cite{liu2025flow} is a popular RL method that extends GRPO from the text domain~\cite{shao24deepseekmath,guo2025deepseek} to image generation using diffusion models, showing notable gains. However, it only uses rewards from the final step of generation, overlooking the denoising trajectory across steps. To overcome this, we propose Stepwise-Flow-GRPO, which introduces step-wise rewards throughout the diffusion process, resulting in more fine-grained guidance and superior performance in image-text alignment and generation quality.

\noindent \textbf{Concurrent works.} TempFlow-GRPO~\cite{he2025tempflow} and Granular-GRPO~\cite{zhou2025fine} also address the credit assignment limitation in Flow-GRPO. While both works recognize that early denoising steps have an outsized impact on final quality, they approach the problem differently: TempFlow-GRPO applies hand-designed noise-level weighting to final-image advantages, while Granular-GRPO optimizes only the first half of steps. We optimize telescoping reward gains that are data-driven and require no manual scheduling or step selection. We provide a detailed comparison in \cref{app:concurrent_work}.

\section{Background}
\label{sec:background}

\noindent \textbf{Flow-matching models} learn continuous-time normalizing flows by training on interpolated data. Given data $x_0 \sim p_\text{data}$ and noise $x_1 \sim \calN(0,I)$, rectified flow \cite{liu2022flow} defines $x_t = (1-t) x_0 + t x_1$ for $t \in [0,1]$. The model $v_\theta(x_t, t, c)$ is trained to predict the velocity field $\dot x_t = x_1 - x_0$ via
\begin{align*}
    \calL(\theta) &= \E_{t,x_0,x_1}[||\dot x_t - v_\theta(x_t,t,c)||^2]
\end{align*}
where $c$ is the prompt for generating $x_0$. Generation follows the ODE: $dx_t = v_\theta(x_t,t,c)dt$. Starting from noise $x_1\sim\calN(0,I)$ at $t=1$ and integrating backward in time to $t=0$ produces samples from the learned distribution.

\noindent \textbf{Flow-GRPO} enables RL fine-tuning of flow models by converting the deterministic ODE into a stochastic SDE. The key insight from \citet{liu2022flow} is the construction of an SDE with matched marginals to the ODE based on the Fokker-Planck equation. The SDE is given by
\begin{align}
    dx_t &= \left[v_t(x_t) \pm \frac{\sigma_t^2}{2t}\hat x_1\right]dt + \sigma_t dw_t \label{eq:sde}
\end{align}
where $\hat x_1 = x_t + (1-t)v_t(x_t)$ is the predicted noise, $\sigma_t$ is an arbitrary noise schedule, $v_t(x_t):=v_\theta(x_t,t,c)$, and $dw_t$ is Brownian motion. In the $\pm$ operator, $+$ corresponds to the forward process (corruption with noise) and $-$ to the reverse process (denoising), obtained via Anderson's time reversal formula~\cite{anderson1982reverse}. This SDE provides stochasticity for RL exploration while preserving the marginals of the original flow. They solve it using the Euler-Maruyama discretization
\begin{align}
    x_{t-\Delta t} &:= x_{t} - \left[v_t(x_t) - \frac{\sigma_t^2}{2t}\hat x_1\right]\Delta t + \sigma_t \sqrt{\Delta t }\epsilon \label{eq:euler_maruyama}
\end{align}
where $\epsilon \sim \calN(0,I)$ injects the stochasticity.
Therefore, for any step $t$, the marginal probability for the previous step $t - \Delta t$ is given by $\pi_\theta(x_{t-\Delta t} | x_t) =$
\begin{align}
    \calN\left(x_{t-\Delta t} ; x_t -\left[v_t(x_t)-\frac{\sigma_t^2}{2t}\hat x_1\right]\Delta t, \sigma_t^2 \Delta t I\right) \label{eq:marginal}
\end{align}
In the rest of the paper, we use $t \in \{T,T-1,\ldots,1,0\}$ to index discrete steps in the RL objective, whereas the earlier $t \in [0,1]$ denotes continuous time in the SDE. This notational convenience simplifies our exposition and should be clear from context. The two are related by $t_{\text{cont.}} = \tfrac{t_{\text{disc.}}}{T}$ and $\Delta t_{\text{disc.}} = \tfrac{1}{T}$.

\citet{liu2025flow} use GRPO~\cite{shao24deepseekmath} to optimize the policy $\pi_\theta$ using group-relative advantages. For each prompt $c$, they generate $N$ trajectories $(x_T^i,\ldots,x_0^i)_{i\in[N]}$ and reward them on the \emph{final image in each trajectory only}, $r_i=R(x_0^i,c)$. We denote by $R(x,c)$ the reward of image $x$ given prompt $c$, and experiment with several different reward functions in~\cref{sec:experiments}. The advantage of trajectory $i$ is
\begin{align*}
  A_i
  = \frac{r_i - \mean}{\std}
\end{align*}
where
\begin{align*}
  \mean
  = \frac{1}{N} \sum_{j=1}^N r_j\,, \quad
  \std
  = \sqrt{\frac{1}{N} \sum_{j = 1}^N (r_j - \mean)^2}
\end{align*}
The policy is updated to locally maximize $J(\theta) =$
\begin{align}
  \frac{1}{N T} \sum_{i=1}^N \sum_{t=0}^{T-1}
  \ell(\rho_t^i(\theta), A_i) - \beta \KL^{i, t}(\pi_\theta || \pi_\tref)
  \label{eq:flow-grpo}
\end{align}
where
\begin{align*}
  \ell(\rho_t^i(\theta), A_i)
  = \min(\rho_t^i(\theta)A_i, \clip(\rho_t^i(\theta), 1-\eps, 1+\eps)A_i)
\end{align*}
is the advantage of trajectory $i$ weighted by the minimum of the unclipped and clipped propensity ratios at step $t$
\begin{align*}
  \rho_t^i(\theta)
  = \tfrac{\pi_\theta(x_t^i|x_{t+1}^i, c)}{\pi_{\text{old}}(x_t^i|x_{t+1}^i, c)}
\end{align*}
between the current policy $\pi_\theta$ in~\cref{eq:marginal} and sampling policy $\pi_\text{old}$,
\begin{align*}
  \clip(x,1-\epsilon,1+\epsilon)
  = \max(1-\epsilon, \min(1+\epsilon, x))
\end{align*}
is a clipping function parameterized by $\epsilon \in [0, 1]$, and
\begin{align*}
  \KL^{i, t}(\pi_\theta || \pi_\tref)
  = \KL(\pi_\theta(x_t^i | x_{t+1}^i, c)||\pi_\tref(x_t^i | x_{t+1}^i, c))
\end{align*}
is the KL penalty between the current policy $\pi_\theta$ and initial policy $\pi_\tref$, at step $t$ of trajectory $i$.

GRPO combines two mechanisms to stabilize policy gradient optimization: propensity ratio clipping and KL regularization. The clipping $\clip(\rho_t^i(\theta), 1-\eps, 1+\eps)$ bounds the propensity ratio to prevent destructively large policy updates. The KL penalty $\beta \KL^{i, t}(\pi_\theta||\pi_\tref)$ anchors the learned policy at the initial policy $\pi_\tref$, preserving its general capabilities while adapting to the reward function. Together, these create a trust region that limits how far the policy can deviate from both the sampling policy (clipping with $\pi_\text{old}$) and its initialization (KL with $\pi_\tref$)~\cite{schulman15trust}.

In the on-policy setting, when $\pi_\text{old} = \pi_\theta$ at all iterations, the importance ratio becomes $\rho_t^i(\theta) = 1$, reducing GRPO to KL-regularized policy gradient regardless of $\eps$ \citep{todorov06linearlysolvable}. This simplified variant is commonly used when trajectories can be generated efficiently, as it removes both the clipping hyperparameter $\eps$ and the complexity of importance weighting, while retaining gradient stabilization through the KL term.

\section{Problems with Uniform Credit Assignment}
\label{sec:key insight}

Flow-GRPO uses the same advantage $A_i$---calculated on the final image $x_0^i$---for every denoising step, treating all steps as equally responsible for final quality. This uniform credit assignment suffers from two fundamental issues that stem from ignoring the temporal structure of diffusion generation.

\noindent \textbf{Problem 1: Ignoring hierarchical frequency structure.} 
The diffusion process has inherent temporal hierarchicy: early steps establish composition and layout (low-frequency structure), while later steps refine details and textures (high-frequency structure). 
This hierarchy arises because different spatial frequencies become distinguishable from noise at different times during denoising.
To see why, consider the rectified flow interpolation in frequency space: $\tilde x_t(k) = (1-t)\tilde x_0(k) + t \tilde x_1(k)$ for spatial frequency $k$. Natural images concentrate energy at low frequencies with power spectrum $\propto |k|^{-\alpha}$ for $\alpha > 0$~\cite{field1987relations,ruderman1994statistics}, while Gaussian noise $\tilde x_1\sim\calN(0,I)$ has flat power across all frequencies. The signal-to-noise ratio at frequency $k$ and time $t$ is therefore
\begin{align*}
    \SNR_t(k)=\left(\frac{1-t}{t}\right)^2 \frac{1}{|k|^\alpha}.
\end{align*}
The $1/|k|^\alpha$ factor means low frequencies always have higher SNR than high frequencies.
As denoising proceeds $(t\to 0)$, the global prefactor $((1-t)/t)^2$ grows, progressively lifting higher frequencies above the noise floor.
The result is a coarse-to-fine generation order: at $t\approx 1$, only low-frequency information is recoverable; fine details emerge only near $t=0$. Uniform credit assignment ignores this structure, treating a step that determines object layout identically to one that sharpens an edge.

\noindent \textbf{Problem 2: Rewarding mistakes that get corrected.} A trajectory might make a mistake at $t\approx 1$ (like incorrect object color) that later steps correct. If the final image has a high reward, Flow-GRPO reinforces \emph{all} steps---including the early mistakes---potentially exacerbating artifacts and compositional errors that are corrected later. \cref{fig:teaser} shows this on a real-world example: reward trajectories are non-monotonic, with intermediate dips that later recover, yet uniform credit assignment treats all steps equally.

\section{Stepwise-Flow-GRPO}
\label{sec:stepwise-flow-grpo}

To address these problems, we introduce stepwise credit assignment through two key ideas: (1) estimating intermediate rewards by predicting clean images from noisy states, and (2) computing advantages based on per-step reward improvements rather than the final reward alone. We also introduce a novel SDE that produces higher-quality denoised images. We now describe each component.

\subsection{Stepwise Rewards}

We cannot directly evaluate reward models on noisy intermediate states $x_t$, but we can estimate the underlying clean image and reward that prediction. The Tweedie formula~\cite{efron2011tweedie} provides $\hat x_0(t) := \E[x_0|x_t] = x_t - t\hat x_1$, where $\hat x_1$ is the predicted noise already computed at each step in the generation process in \cref{eq:euler_maruyama}. This makes the one-step Tweedie estimate essentially free. To get higher-quality estimates, we denoise from $x_t$ toward $x_0$ over multiple steps. We discretize the interval $[t, 0]$ into $T'$ steps and solve the flow ODE using forward Euler: starting from $x_t$, we iteratively apply the flow for steps $t, t(T'-1)/T', \ldots, t/T', 0$. When $T'=1$, this reduces to the one-step Tweedie estimate. In practice, we find that $T'=5$ ODE substeps provide strong reward signal with minimal additional cost.

We compute \textbf{stepwise rewards} $r_t^i = R(\hat x_0^i(t), c)$ at each step $t$ of trajectory $i$, allowing us to evaluate how intermediate states progress toward high-quality images rather than relying solely on final outcomes. Crucially, each $r_t^i$ is estimated using a single deterministic denoising trajectory of $T'$ substeps starting from $x_t^i$---no averaging over multiple samples is needed. Since the denoising from each state $x_t^i$ is independent, all $T$ reward estimates for a trajectory can be computed in parallel (see \cref{app:timing}).

\subsection{Stepwise Policy Optimization}
\label{subsec:stepwise_policy_optimization}

\begin{figure}[t]
    \centering
    \includegraphics[width=\linewidth]{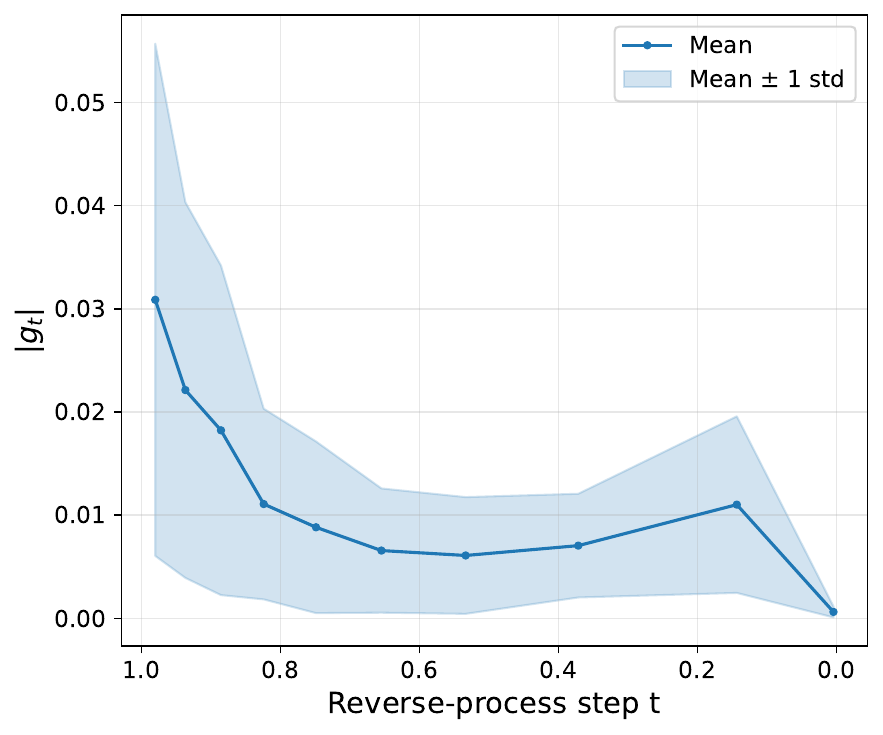}
    \caption{\textbf{Gain magnitudes across steps.} Mean absolute gain $\E_i[|g_t^i|]$ measured on 256 GenEval prompts using PickScore. Early steps show larger gains, indicating that compositional decisions drive most reward improvement.}
    \label{fig:gains}
\end{figure}

We do not optimize stepwise rewards directly, as this would optimize high-scoring intermediary Tweedie estimates instead of a high-scoring final image. Instead, we compute \textbf{stepwise gains} $g_t^i := r_{t - 1}^i - r_t^i$, which measure each step's marginal reward improvement. Steps that increase the reward receive positive reinforcement while steps that decrease it are penalized.

Crucially, these gains telescope $\sum_{t=1}^T g_t^i = r_0^i - r_T^i$, meaning that maximizing stepwise gains is equivalent to maximizing the improvement from initial noise $x_T$ to final image $x_0$ for a fixed initial condition. This connects local per-step optimization to the global objective of maximizing final reward. As shown in \cref{fig:gains}, the magnitudes of the gains are initially large and diminish as $t \to 0$, suggesting that early compositional decisions drive most of the reward improvement. Under certain submodularity assumptions, greedy maximization of such gains would be provably near-optimal (\cref{sec:submodular optimization}).

\noindent \textbf{Joint normalization preserves temporal structure.} Following GRPO, we transform gains into group-relative advantages. A key design choice is whether to normalize gains separately at each step or jointly across all steps. We choose joint normalization to preserve the larger magnitudes of early gains, since per-step normalization would artificially inflate noise in later steps where reward improvements are small. The group-relative advantage at step $t$ of trajectory $i$ is
\begin{align}
  \tilde{A}_t^i
  = \frac{g_t^i - \mean}{\std}
  \label{eq:stepwise-flow-grpo advantage}
\end{align}
where
\begin{align*}
  \mean
  = \frac{1}{NT} \sum_{j, k} g_k^j, \quad
  \std
  = \sqrt{\frac{1}{NT} \sum_{j, k} (g_k^j - \mean)^2}
\end{align*}
for $j \in \{1, \dots, N\}$ and $k \in \{0, \dots, T - 1\}$. We update the policy to maximize $J(\theta) =$
\begin{align}
  \frac{1}{NT} \sum_{i=1}^N \sum_{t=0}^{T-1}
  \left[\ell(\rho_t^i(\theta), \tilde{A}_t^i) - \beta \KL^{i,t}(\pi_\theta || \pi_\tref)\right]
  \label{eq:stepwise-flow-grpo}
\end{align}
where all the terms are defined as in \cref{eq:flow-grpo}. The main difference from Flow-GRPO is that uniform advantages $A_i$ are replaced with stepwise advantages $\tilde{A}_t^i$. The pseudo-code of our method is in~\cref{alg:stepwise-flow-grpo}. All $N$ trajectories within a group share the same initial noise $x_T$, so that reward differences arise solely from the stochastic denoising process.

\begin{algorithm}[t!]
\caption{Stepwise-Flow-GRPO}
\label{alg:stepwise-flow-grpo}
\begin{algorithmic}
\Require Base policy $\pi_{\text{ref}}$, prompts $\mathcal{P}$, number of steps $T$, batch size $N$, number of substeps $T'$
\Ensure Initialize RL policy $\pi_\theta \gets \pi_\tref$
\While{not converged}
    \State Sample prompt $c \sim \mathcal{P}$
    \State Initialize $x_T\sim \calN(0,I)$
    \For{$i = 1, \ldots, N$}
        \State $x_T^i \gets x_T$ \Comment{All trajectories share initial noise}
        \State Generate trajectories $(x_{T - 1}^i, \ldots, x_0^i) \sim \pi_\theta(\cdot|c)$ autoregressively using~\cref{eq:euler_maruyama}
        \State Compute $\hat{x}_0^i(t)$ using $T'$ substeps for all $t$
        \State Compute rewards $r_t^i \gets R_t(\hat{x}_0^i(t), c)$ for all $t$
        \State Compute gains $g_t^i \gets r_{t - 1}^i - r_t^i$ for all $t$
    \EndFor
    \State Compute $\tilde A_t^i$ in \cref{eq:stepwise-flow-grpo advantage} for all $t$ and $i$
    \State Update $\theta$ by policy gradient using an AdamW step on loss $-J(\theta)$ in \cref{eq:stepwise-flow-grpo}
\EndWhile
\State \Return $\pi_\theta$
\end{algorithmic}
\end{algorithm}

\subsection{Connection to Submodular Optimization}
\label{sec:submodular optimization}

Our greedy gain maximization has theoretical grounding in recent adaptive submodular optimization results. Consider a simplified on-policy variant of our method where $\pi_\text{old} = \pi_\theta$ (so $\rho_t^i(\theta) = 1$) and advantages are replaced by raw gains $g_t^i$. Then the objective reduces to
\begin{align}
  J(\theta) = \frac{1}{NT} \sum_{i=1}^N \sum_{t=1}^T g_t^i - \beta \KL(\pi_\theta || \pi_\tref)
  \label{eq:online_policy_gradient}
\end{align}
This formula is algebraically equivalent to the adaptive policy gradient objective in \citet{kveton25adaptive}. \citet{kveton25adaptive} showed that greedy KL-regularized policy gradients can learn near-optimal policies when reward gains are monotone and submodular—analogous to classic guarantees for greedy submodular maximization~\citep{nemhauser78approximation,golovin11adaptive}.

The key difference from the classic policy gradient is that rather than optimizing the total trajectory reward (as in Flow-GRPO), we optimize per-step gains. Our setting generalizes \citet{kveton25adaptive} by allowing off-policy optimization ($\pi_\text{old} \neq \pi_\theta$) and group-relative advantages, and represents the first application of adaptive gain maximization to GRPO and flow models. While we do not formally verify submodularity of our reward functions, our empirical successes and diminishing gains in \cref{fig:gains} suggest that a submodularity-like structure is present.

\subsection{Design Variations}

In \cref{app:variations}, we explore several design alternatives: (1) centering gains with an exponential moving average baseline to reduce temporal variance, (2) generalized advantage estimation (GAE) to let each step receive partial credit for future gains it enables, trading off bias and variance via exponential discounting, (3) an ODE-based formulation using progressive distillation between successive Tweedie estimates $\hat x_0(t)$ and $\hat x_0(t-2\Delta t)$. While these offer different computational tradeoffs, the gain we presented in~\cref{subsec:stepwise_policy_optimization} performs best in our experiments.

\subsection{Improved Sampling via DDIM-Style Updates}
\label{subsec:improved_sde}

The SDE in~\cref{eq:sde} enables RL by injecting the stochasticity needed for exploration, while matching the marginals to the ODE ensures that optimizing the SDE policy also improves the deterministic ODE used at inference.

However, we find empirically that the noise injected by the SDE produces visually degraded samples compared to deterministic ODE integration. Since reward models are trained on clean images, this degrades the reward estimates and slows down optimization. We adopt a DDIM-style sampling strategy~\cite{song2021denoising} that produces higher-quality samples for the reward model while preserving the stochasticity required for policy gradients.

The DDIM update rule interpolates between deterministic and stochastic sampling
\begin{align*}
   x_{t-\Delta t} = \sqrt{\alpha_{t-\Delta t}}\,\hat x_0(t) + \sqrt{\beta_{t-\Delta t} - \sigma_t^2}\,\hat x_1 + \sigma_t \epsilon
    \label{eq:ddim_update}
\end{align*}
where $\sqrt{\alpha_{t-\Delta t}}$ is the signal strength, $\sqrt{\beta_{t-\Delta t}}$ is the noise strength, and $\sigma_t$ controls stochasticity, To match the marginals of rectified flow $x_t = (1-t)x_0 + tx_1$, we set $\alpha_t = (1-t)^2$ and $\beta_t = t^2$, yielding $x_{t-\Delta t} =$
\begin{align}
   (1-(t-\Delta t))\,\hat x_0(t) + \sqrt{(t-\Delta t)^2 - \sigma_t^2}\,\hat x_1 + \sigma_t \epsilon
\end{align}
When $\sigma_t = 0$, this recovers the deterministic flow ODE.

Exact marginal matching would require the coefficient of $\hat x_1$ to be $(t-\Delta t) - \tfrac{\sigma_t^2}{(2t)}$ (derived in \cref{app:improved_sde}). The DDIM form $\sqrt{(t-\Delta t)^2 - \sigma_t^2} = (t-\Delta t) - \tfrac{\sigma_t^2}{2(t-\Delta t)} + O(\sigma_t^4)$ is a close approximation for small $\sigma_t$ and $\Delta t$, and produces higher-quality images in practice.

We use $\sigma_t = \eta(t-\Delta t)\sqrt{1-t}$, where $\eta$ controls exploration strength. This schedule injects maximum stochasticity at $t=1$ (heavy noise), smoothly anneals to $\sigma_t \to 0$ as $t\to 0$ to recover the deterministic ODE, and maintains small $\sigma_t$ throughout to match the marginal. The result is a Gaussian policy $\pi_\theta(x_{t-1}|x_t,c) = \mathcal{N}(x_{t-1}; \mu_t, \sigma_t^2 I)$ with mean
\begin{align}
    \mu_t = (1-(t-\Delta t))\,\hat x_0(t) + \sqrt{(t-\Delta t)^2 - \sigma_t^2}\,\hat x_1
\end{align}
This allows computing importance ratios $\rho_t^i(\theta)$ and KL penalties in~\cref{eq:flow-grpo}. This approach significantly improves visual quality while preserving the ability to optimize via GRPO. A similar result was derived in a recent concurrent work~\cite{wang2025coefficientspreserving}.

\begin{figure}[t]
    \centering
    \includegraphics[width=\linewidth]{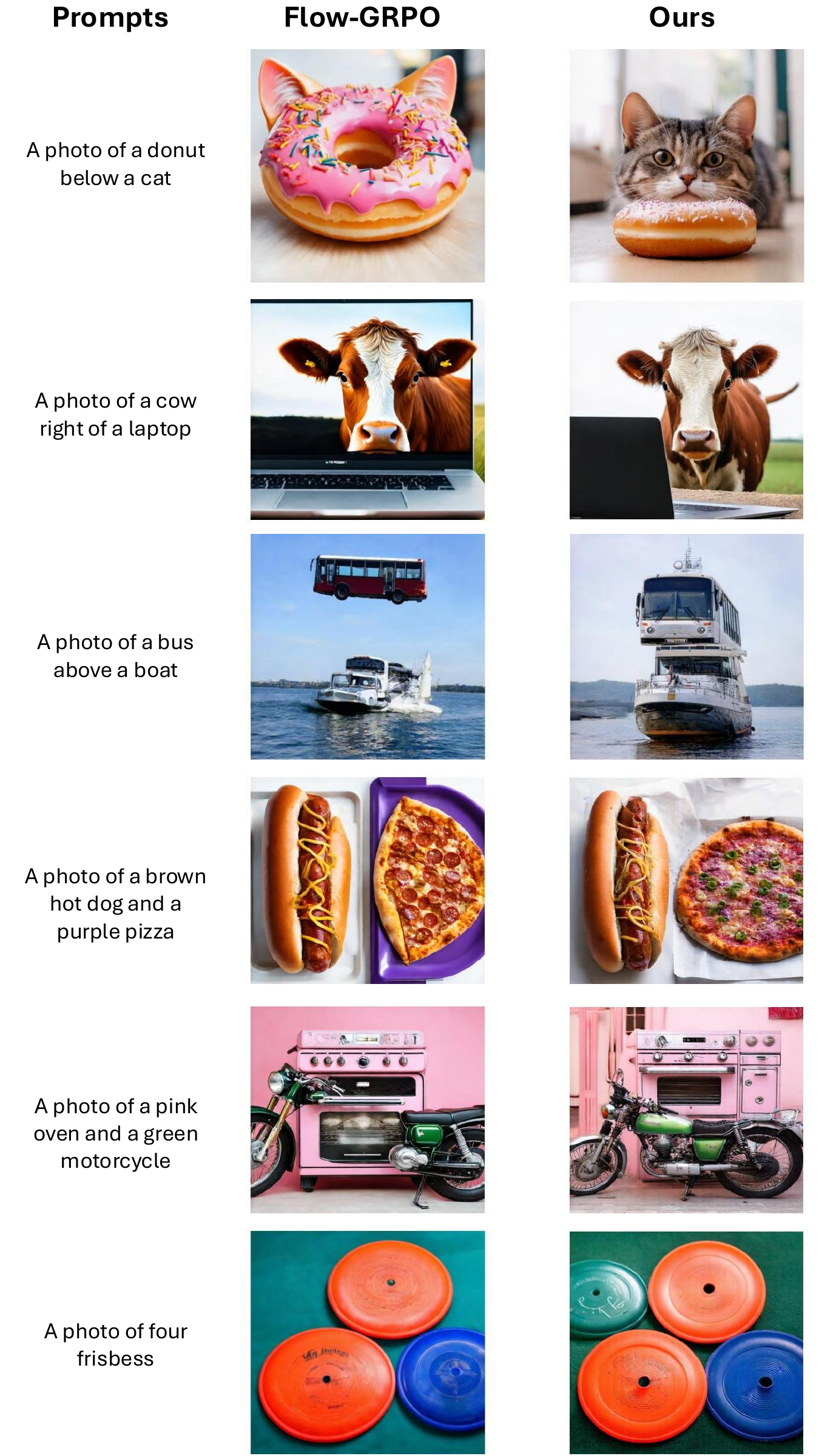}
    \caption{\textbf{Qualitative results}. We compare our Stepwise-Flow-GRPO with Flow-GRPO and observe better spatial reasoning, attribute binding, and counting performance.}
    \label{fig:qual_results}
\end{figure}

\begin{figure*}[t!]
    \centering
    \begin{subfigure}[t]{0.24\textwidth}
        \includegraphics[width=\linewidth]{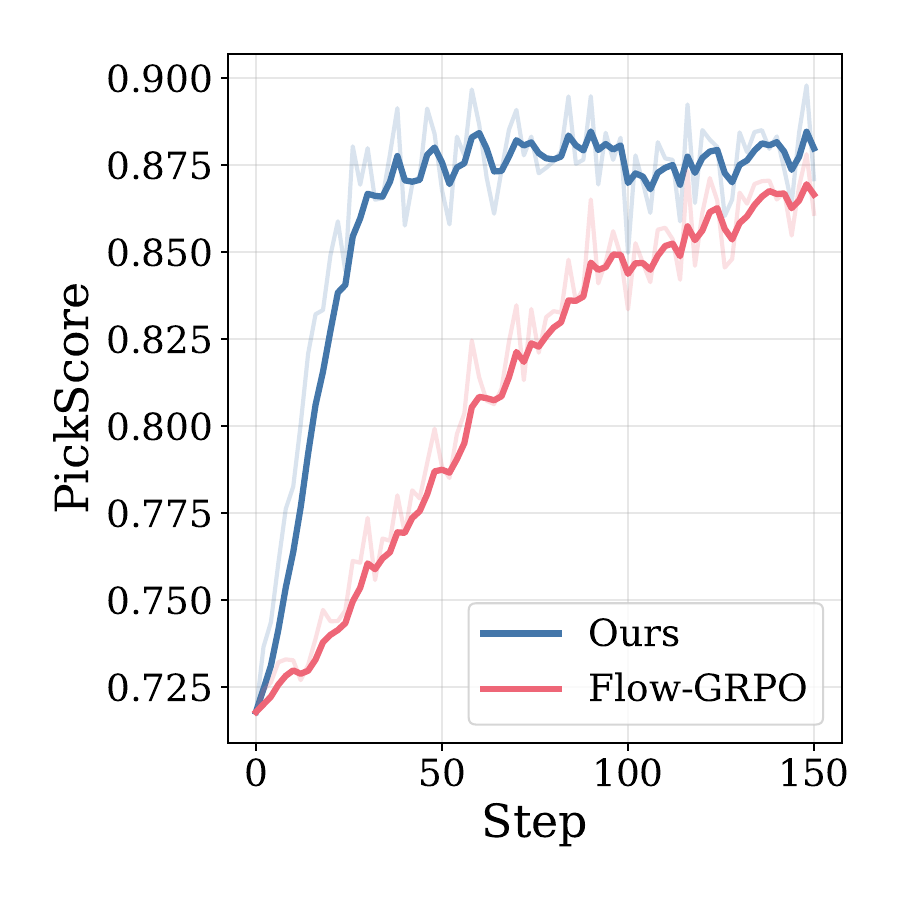}
        \caption{Pickscore (GenEval)}
    \end{subfigure}
    \hfill
    \begin{subfigure}[t]{0.24\textwidth}
        \includegraphics[width=\linewidth]{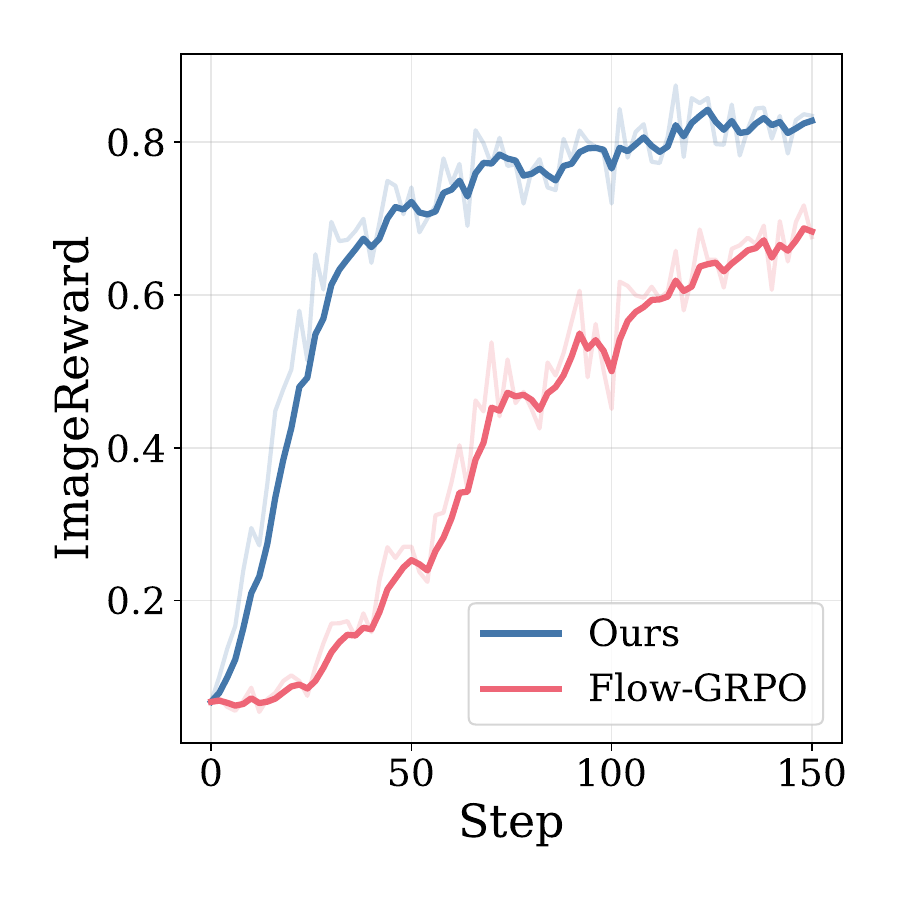}
        \caption{ImageReward (GenEval)}
    \end{subfigure}
    \hfill
    \begin{subfigure}[t]{0.24\textwidth}
        \includegraphics[width=\linewidth]{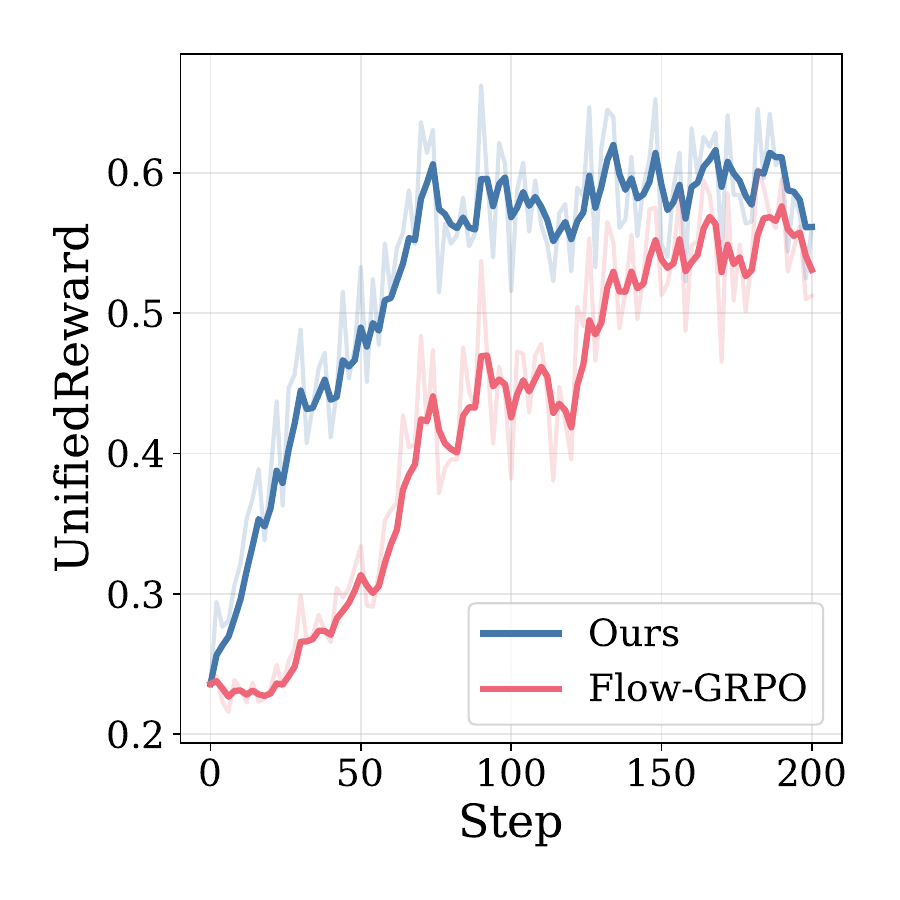}
        \caption{UnifiedReward (GenEval)}
    \end{subfigure}
    \begin{subfigure}[t]{0.24\textwidth}
        \includegraphics[width=\linewidth]{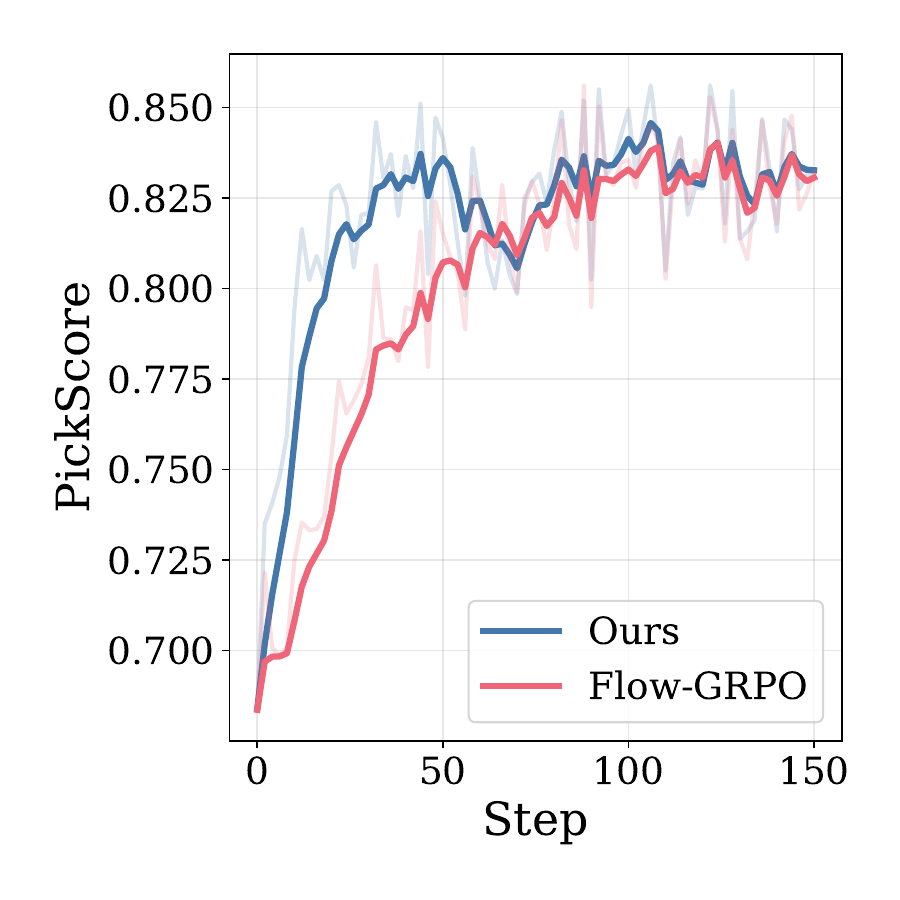}
        \caption{PickScore (PickScore Dataset)}
    \end{subfigure}
    \caption{\textbf{Sample efficiency across reward functions.} Stepwise-Flow-GRPO consistently outperforms Flow-GRPO in reward per training step across all settings, achieving both faster convergence and superior final performance in 3 out of 4 settings.}
    \label{fig:flowsde_results}
\end{figure*}

\begin{figure*}[t!]
    \centering
    \begin{subfigure}[t]{0.24\textwidth}
        \includegraphics[width=\linewidth]{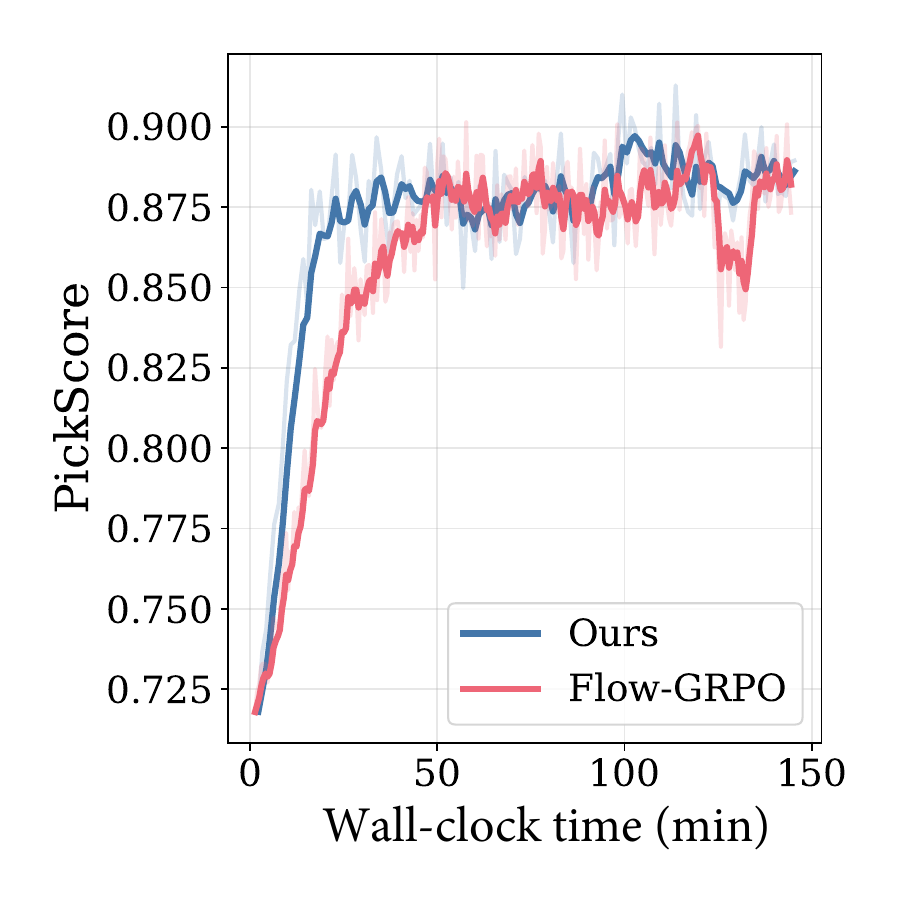}
        \caption{Pickscore (GenEval)}
    \end{subfigure}
    \hfill
    \begin{subfigure}[t]{0.24\textwidth}
        \includegraphics[width=\linewidth]{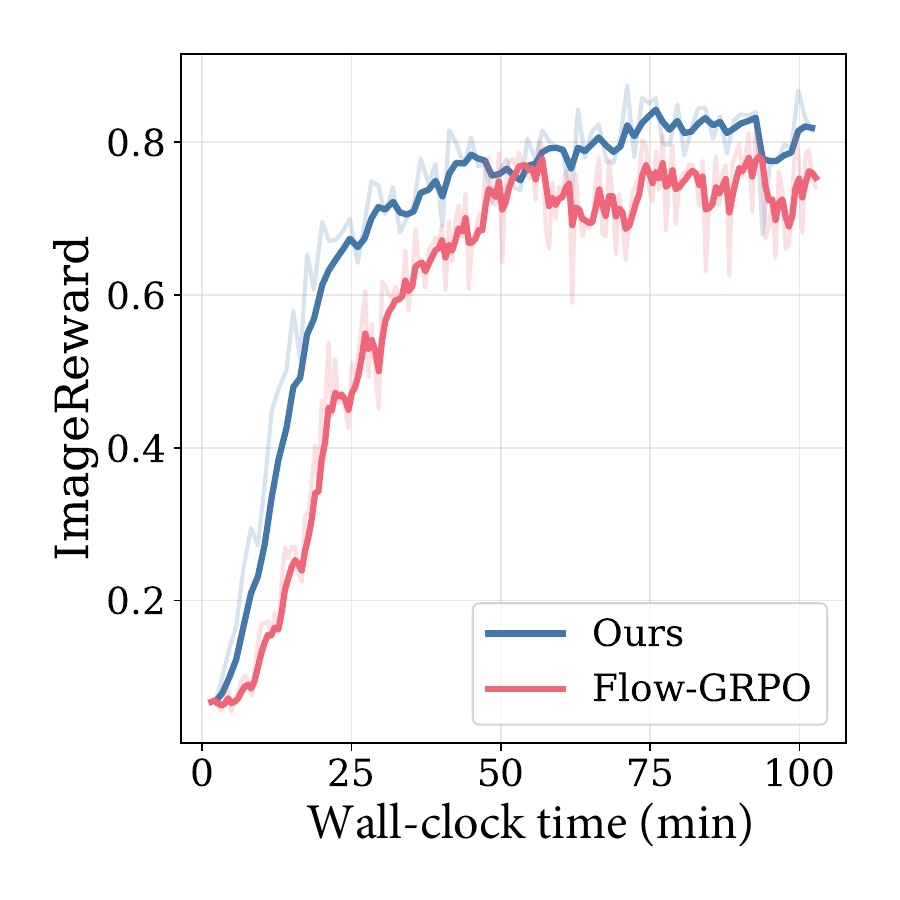}
        \caption{ImageReward (GenEval)}
    \end{subfigure}
    \hfill
    \begin{subfigure}[t]{0.24\textwidth}
        \includegraphics[width=\linewidth]{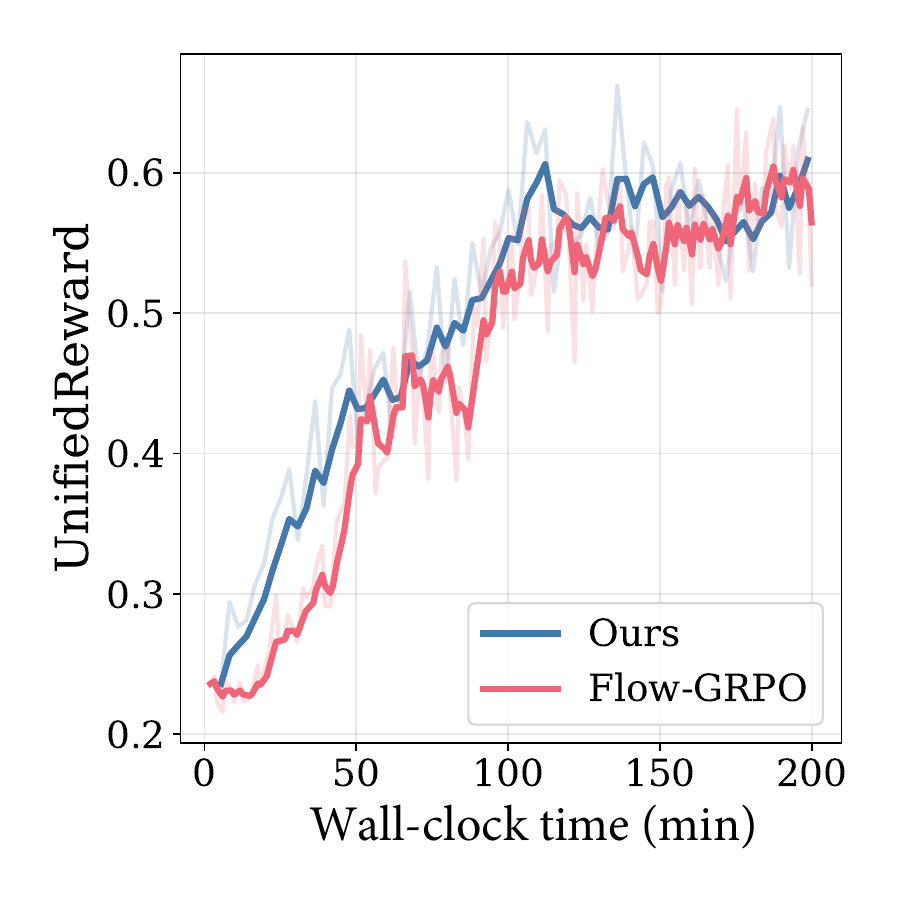}
        \caption{UnifiedReward (GenEval)}
    \end{subfigure}
    \hfill
    \begin{subfigure}[t]{0.24\textwidth}
        \includegraphics[width=\linewidth]{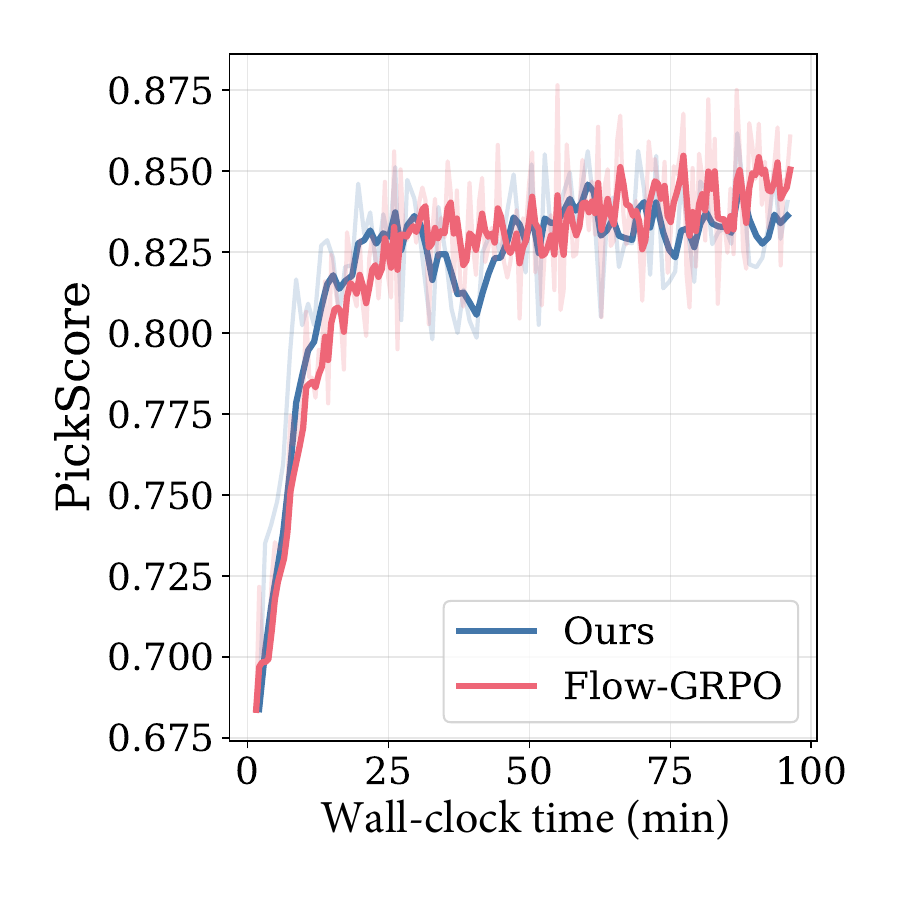}
        \caption{Pickscore (PickScore Dataset)}
    \end{subfigure}
    \caption{\textbf{Wall-clock efficiency matches sample efficiency gains.} Reward versus wall-clock time for the same settings as \cref{fig:flowsde_results}. Despite additional computational cost for intermediate denoising, Stepwise-Flow-GRPO converges faster in wall-clock time, achieving visibly superior performance in 3 out of 4 settings.}
    \label{fig:wallclock}
\end{figure*}

\begin{figure*}[t!]
    \centering
    \hfill
    \begin{subfigure}[t]{0.24\textwidth}
        \includegraphics[width=\linewidth]{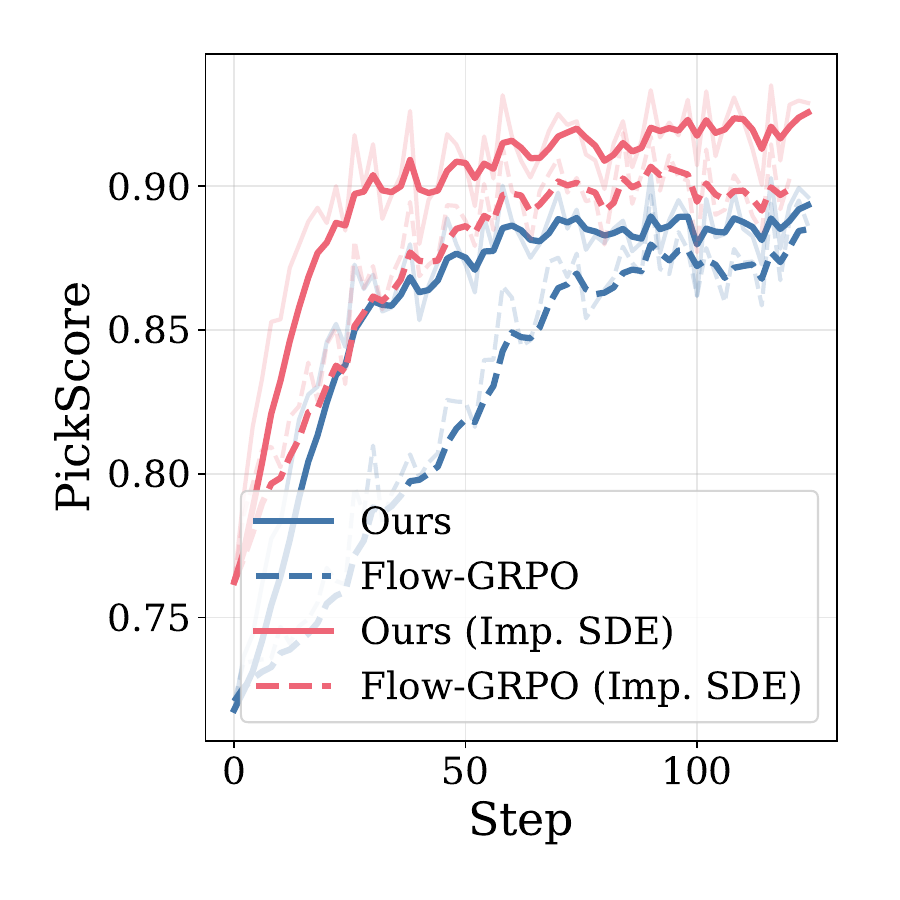}
        \caption{PickScore (GenEval)}
    \end{subfigure}
    \hfill
    \begin{subfigure}[t]{0.24\textwidth}
        \includegraphics[width=\linewidth]{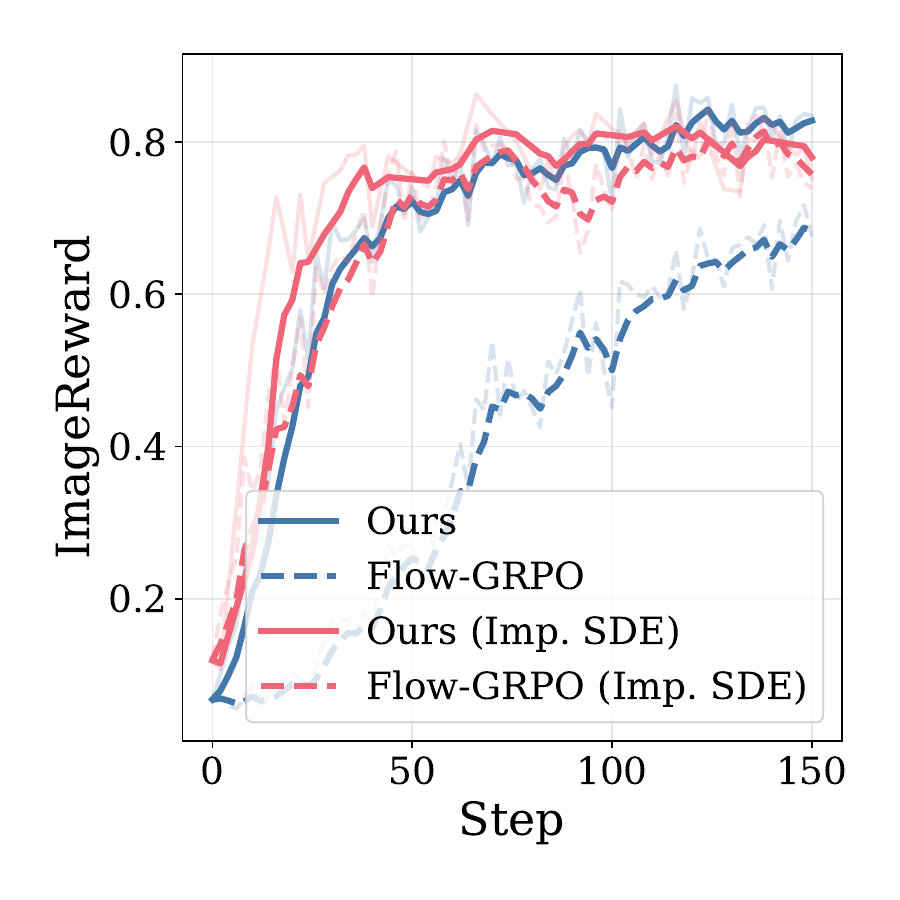}
        \caption{ImageReward (GenEval)}
    \end{subfigure}
    \hfill
    \begin{subfigure}[t]{0.24\textwidth}
        \includegraphics[width=\linewidth]{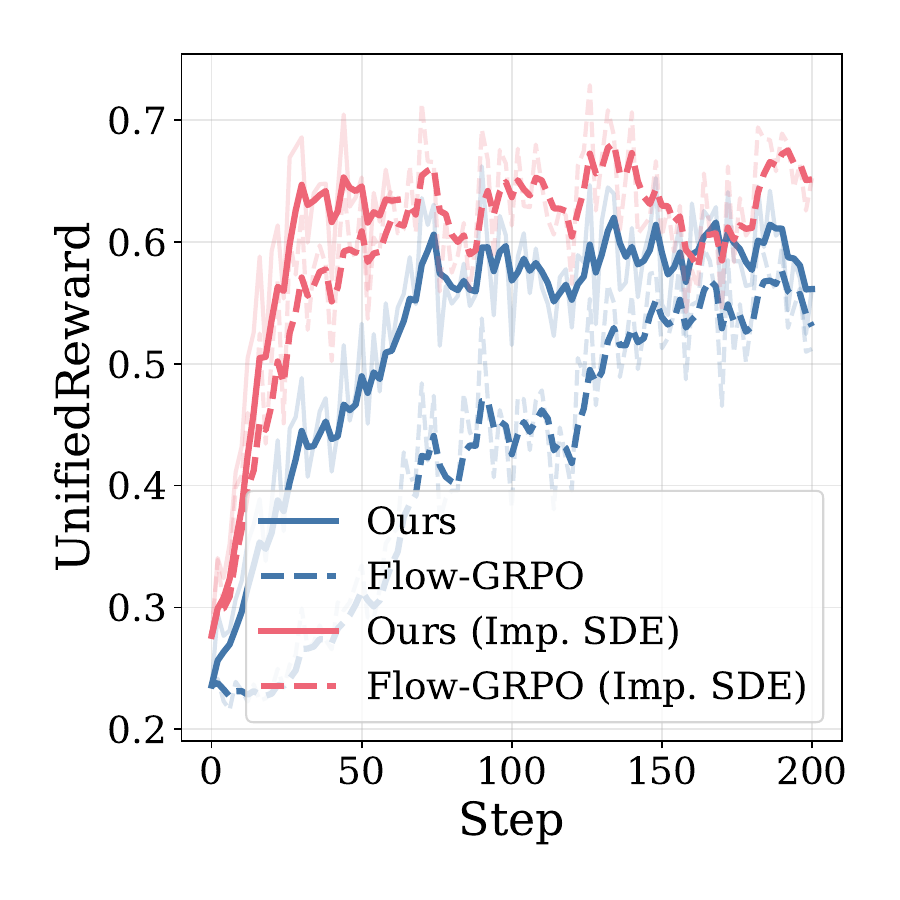}
        \caption{UnifiedReward (GenEval)}
    \end{subfigure}
    \hfill
    \begin{subfigure}[t]{0.24\textwidth}
        \includegraphics[width=\linewidth]{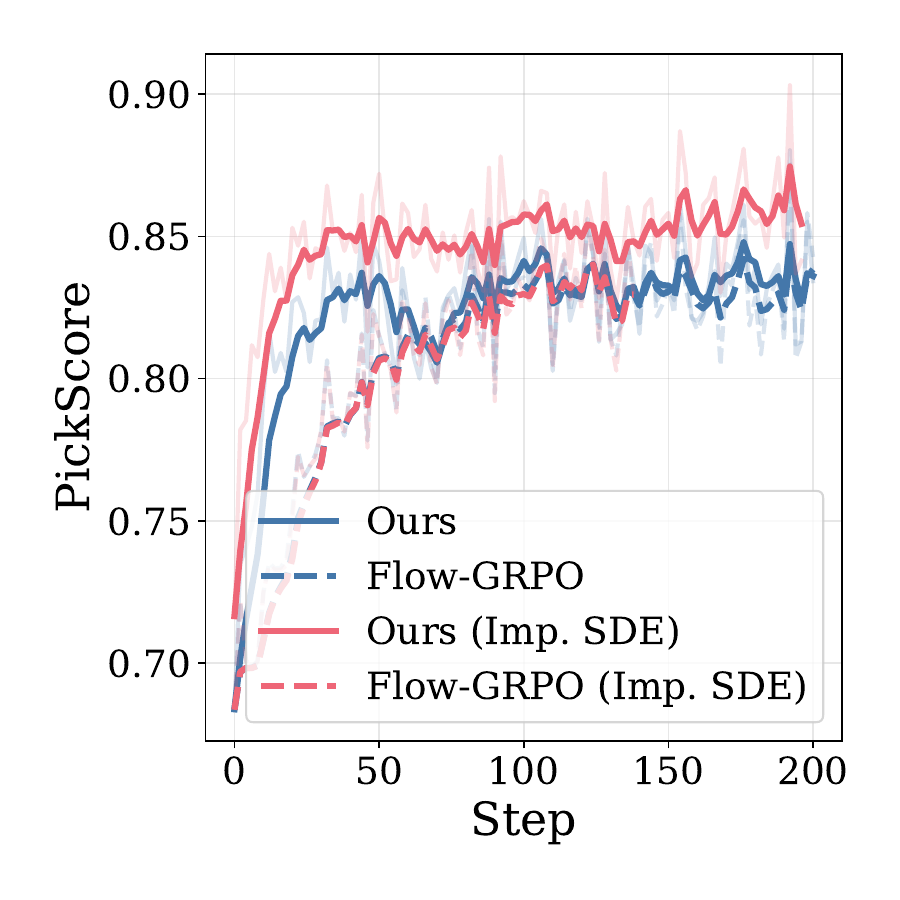}
        \caption{Pickscore (PickScore Dataset)}
    \end{subfigure}
    \caption{\textbf{Stepwise credit assignment remains effective with improved SDE.} Reward versus training step when both methods use the DDIM-inspired SDE from \cref{subsec:improved_sde}. Stepwise-Flow-GRPO retains its sample efficiency advantage, demonstrating that the improvements in credit assignment and sampling are complementary.}
    \label{fig:improved_sde}
\end{figure*}

\begin{table*}[!t]
\small
\centering
\begin{tabular}{l | c | c c c c c c}
\toprule
\textbf{Experiment} & \textbf{Overall} & \textbf{Single Obj.} & \textbf{Two Objs.} & \textbf{Counting} & \textbf{Colors} & \textbf{Position} & \textbf{Attr. Binding} \\
\midrule
\multicolumn{8}{c}{\textit{Pretrained Models}} \\
\midrule
SD3.5-M (cfg=1.0) & 0.28 & 0.71 & 0.23 & 0.15 & 0.45 & 0.05 & 0.08 \\
SD3.5-M (cfg=4.5) & 0.63 & 0.98 & 0.78 & 0.50 & 0.81 & 0.24 & 0.52 \\
\midrule
\multicolumn{8}{c}{\textit{Fine-tuned Models}} \\
\midrule
SD3.5-M+Flow-GRPO (cfg=1.0 PickScore) & 0.60 & 0.96 & 0.73 & 0.67 & 0.67 & 0.21 & 0.35 \\
\textbf{SD3.5-M+Ours (cfg=1.0 PickScore)} & 0.60 & 0.96 & 0.75 & 0.67 & 0.67 & 0.21 & 0.34 \\
SD3.5-M+Flow-GRPO (cfg=4.5 Pickscore) & 0.68 & \textbf{0.98} & 0.82 & 0.64 & \textbf{0.82} & 0.24 & \textbf{0.59} \\
\textbf{SD3.5-M+Ours (cfg=4.5 PickScore)} & \textbf{0.71} & \textbf{0.98} & \textbf{0.85} & \textbf{0.70} & \textbf{0.82} & \textbf{0.29} & \textbf{0.59} \\
\bottomrule
\end{tabular}
\caption{\textbf{Final model quality on GenEval.} Compositional generation performance for models trained with PickScore reward. Both methods substantially improve over the base model, with our method matching Flow-GRPO at cfg=1.0 and outperforming it across most categories at cfg=4.5, particularly in counting and spatial positioning.}
\label{tab:geneval}
\end{table*}

\section{Experiments}
\label{sec:experiments}

\noindent \textbf{Datasets and metrics.} Our main experiments are conducted on two datasets: PickScore~\cite{pickscore} and GenEval~\cite{ghosh2023geneval}, which tests compositional understanding through spatial relationships, object counts, and attribute binding. We additionally evaluate OCR text rendering in \cref{app:additional_results}. For the final model validation, we evaluate PickScore-trained models using GenEval and provide qualitative examples of their outputs to verify that they match Flow-GRPO without reward hacking or artifacts.

\noindent \textbf{Reward functions.} On GenEval, we train with three pretrained reward models: PickScore~\cite{pickscore}, ImageReward~\cite{imagereward}, and UnifiedReward-7b-v1.5~\cite{unifiedreward}. On the PickScore dataset, we use PickScore as the reward. UnifiedReward requires 8 additional NVIDIA A100 80GB GPUs, deployed via SGLang~\cite{zheng2023efficiently} for efficient inference (prompt in \cref{app:additional_results}). All reward functions use the default Flow-GRPO implementation with efficiency optimizations.

\noindent \textbf{Implementation details.} We use \texttt{SD3.5-Medium}~\cite{stablediffusion35_2024} trained on 8 NVIDIA A100 GPUs, built from the Flow-GRPO codebase~\cite{liu2025flow}. We sample using 10 denoising steps with $\text{cfg}=1.0$~\cite{ho2022classifierfree} (no classifier-free guidance) and batch size 16 to maximize throughput and double training batch sizes. For the SDE formulation, we use $\sigma_t = 0.7\sqrt{t/(1-t)}$ following Flow-GRPO, and additionally evaluate both methods with our improved SDE from~\cref{subsec:improved_sde} using $\eta=0.9$. Our method computes intermediate estimates $\hat{x}_0(t)$ using $T'=5$ flow ODE substeps from each $x_t$. To assess final model quality under standard inference settings, we separately train models with $\text{cfg}=4.5$ and batch size 8 using PickScore on the GenEval dataset, and evaluate with the GenEval benchmark (reward and wall-clock figures in \cref{app:additional_results}).

\subsection{Main Results}

\noindent \textbf{Sample efficiency.} \cref{fig:flowsde_results} shows that Stepwise-Flow-GRPO consistently achieves higher rewards for the final image per training iteration than Flow-GRPO across all reward functions, demonstrating superior sample efficiency. The improvement is most pronounced early in training, where stepwise credit assignment enables faster identification of effective denoising strategies. Our method achieves both faster convergence and superior final performance in 3 out of 4 settings. Importantly, \cref{tab:geneval} shows that despite faster convergence, our method achieves no worse final GenEval accuracy than Flow-GRPO, indicating we improve learning speed without sacrificing final quality.

\noindent \textbf{Wall-clock efficiency.} While Stepwise-Flow-GRPO requires additional computation for intermediate denoising and reward evaluation (timing breakdown in \cref{app:timing}), the improved sample efficiency compensates for this overhead. \cref{fig:wallclock} shows that our method converges faster in wall-clock time, achieving visibly superior performance in 3 out of 4 settings, reaching target reward values with less total computation than Flow-GRPO.

\noindent \textbf{GenEval results.} Our Stepwise-Flow-GRPO matches Flow-GRPO at cfg=1.0 and outperforms it at cfg=4.5 (it is equal or better on all sub-categories) in \cref{tab:geneval}. We compare the methods qualitatively in \cref{fig:qual_results}. The results indicate that our method improves spatial reasoning, attribute binding, and counting capabilities. The generated images also appear to be more realistic and physically plausible. Notably, Flow-GRPO merges objects in the first and fifth rows, and places the bus unrealistically in the sky instead of on the top of the boat in the third row. Additional qualitative examples are given in \cref{app:additional_results}.

\noindent \textbf{Improved SDE results.} \cref{fig:improved_sde} shows that Stepwise-Flow-GRPO has better sample efficiency than Flow-GRPO when both methods use the improved SDE from~\cref{subsec:improved_sde}. While the DDIM-inspired SDE substantially improves both methods, combining it with stepwise credit assignment yields better results than either modification alone, showing that the improvements are complementary.

\subsection{Ablation Studies}

\noindent \textbf{Number of denoising substeps.} We ablate the number of ODE steps $T'$ used to estimate $\hat{x}_0(t)$ in \cref{app:ablations}. When $1 \leq T' \leq 3$, the reward estimates are noisy and degrade training signal, while $6 \leq T' \leq 10$ have negligible quality improvements at increased computational cost. We select $T' = 5$ as the optimal tradeoff between estimate quality and efficiency.

\noindent \textbf{Gain normalization strategy.} We compare joint normalization in \cref{eq:stepwise-flow-grpo advantage} to per-step normalization in \cref{app:ablations}. Joint normalization preserves the temporal structure where early gains are larger (\cref{fig:gains}), allowing optimization to prioritize compositional decisions over detail refinement. Per-step normalization equalizes importance across the steps, resulting in slower convergence (see \cref{app:ablations}).

\section{Conclusions and Future Work}

We introduced Stepwise-Flow-GRPO, which assigns credit to individual denoising
steps using gain-based advantages from intermediate reward estimates, and a
DDIM-inspired SDE that improves reward signal quality. Together, these yield
superior sample efficiency, convergence speed, and training stability over
uniform credit assignment.

Our stepwise gains open several directions: using per-prompt gain variance as a
difficulty signal for curriculum learning, adaptively weighting steps
proportional to their gain variance to focus optimization on high-information
regions of the trajectory, and enabling self-correcting diffusion where models
learn to detect and retry poor intermediate decisions.

\clearpage

{
    \small
    \bibliographystyle{ieeenat_fullname}
    \bibliography{Brano,main}
}


\clearpage
\appendix
\setcounter{page}{1}
\maketitlesupplementary

In this appendix, we cover the following topics: A) our exploration on various design choices B) our improvement on SDE C) ablation studies D) our runtime analysis E) additional results F) comparisons with concurrent work.

\section{Design Variations}
\label{app:variations}

We explored several alternative formulations of stepwise credit assignment to investigate potential improvements in variance reduction and temporal credit propagation. While these variations offer intuitive benefits in principle, our experiments show that the gain formulation from \cref{subsec:stepwise_policy_optimization} performs best in practice.

\subsection{Exponential Moving Average Baseline}

\textbf{Motivation.}
Raw gains $g_t^i = r_{t-1}^i - r_t^i$ can exhibit high variance across training iterations, potentially leading to noisy gradient estimates. A common variance reduction technique in RL is to center advantages using a baseline. We explore using an exponential moving average (EMA) of past gains as a temporal baseline.

\textbf{Formulation.}
We maintain a per-step EMA baseline $b_t$ updated after each training iteration:
\begin{align}
    b_t &\leftarrow \alpha \cdot b_t + (1-\alpha) \cdot \mathbb{E}_i[g_t^i]
\end{align}
where $\alpha = 0.99$ is the decay rate. The centered gains are then:
\begin{align}
    \tilde{g}_t^i = g_t^i - b_t
\end{align}
These centered gains are normalized and converted to advantages as in the standard formulation.

\textbf{Expected benefit.}
By subtracting a running average of typical gain magnitudes at each step, we hoped to reduce variance in advantage estimates while preserving the directional signal about which steps improve reward. This is analogous to value function baselines in policy gradient methods.

\subsection{Generalized Advantage Estimation (GAE)}

\textbf{Motivation.}
Standard gains $g_t^i$ only capture the immediate reward improvement from step $t$. However, a step's true contribution might include its downstream impact on future steps. GAE~\cite{schulman2015high} provides a principled way to trade off bias and variance by exponentially weighting future gains.

\textbf{Formulation.}
We apply GAE to the gain sequence with discount factor $\gamma$:
\begin{align}
    \text{GAE}_t^i = \sum_{k=0}^{T-t} \gamma^k g_{t+k}^i
\end{align}
In practice, we compute this efficiently via:
\begin{align}
    \text{GAE}_t^i = \frac{1}{\gamma^t} \sum_{k=t}^{T} \gamma^k g_k^i
\end{align}
We use $\gamma = 0.95$ following standard RL practice. The GAE values replace raw gains in the advantage calculation.

\textbf{Expected benefit.}
GAE allows early steps to receive credit for setting up conditions that enable large future gains, even if their immediate gain $g_t^i$ is small. For example, a compositional decision at $t \approx T$ might have small immediate reward improvement but enable larger gains at later refinement steps. We hypothesized this could improve credit assignment for temporally extended effects.

\subsection{ODE-Based Progressive Distillation}

\textbf{Motivation.}
The SDE formulation requires introducing stochasticity $\sigma_t$ for policy gradients, which can degrade sample quality. An alternative approach is to avoid the SDE entirely and instead optimize a progressive distillation objective between successive Tweedie estimates.

\textbf{Formulation.}
Instead of computing log-probabilities under a Gaussian policy, we use a progressive distillation loss between the current step's Tweedie estimate $\hat{x}_0(t)$ and the twice-previous step's estimate $\hat{x}_0(t-2\Delta t)$:
\begin{align}
    \mathcal{L}_t^i = -\|\hat{x}_0^i(t-2\Delta t) - \mu_t^i\|^2
\end{align}
where $\mu_t^i$ is the mean of the flow transition from $x_t$ to $x_{t-2\Delta t}$:
\begin{align}
    \mu_t^i = \hat{x}_0^i(t) \cdot (1-(t-2\Delta t)) + \hat{x}_1^i(t) \cdot (t-2\Delta t)
\end{align}
We then scale this loss by gain-based advantages as in standard GRPO.

\textbf{Expected benefit.}
This formulation completely avoids the stochasticity required for the SDE, potentially improving sample quality. The distillation objective encourages consistency between successive predictions of the final image, with stronger enforcement on steps with high gains. We hypothesized this could provide cleaner gradients while maintaining stepwise credit assignment.

\subsection{Experimental Results}
We evaluate all variations on the PickScore reward using the GenEval dataset, following the same training protocol as our main experiments.

\begin{figure}[t]
    \centering
    \includegraphics[width=\linewidth]{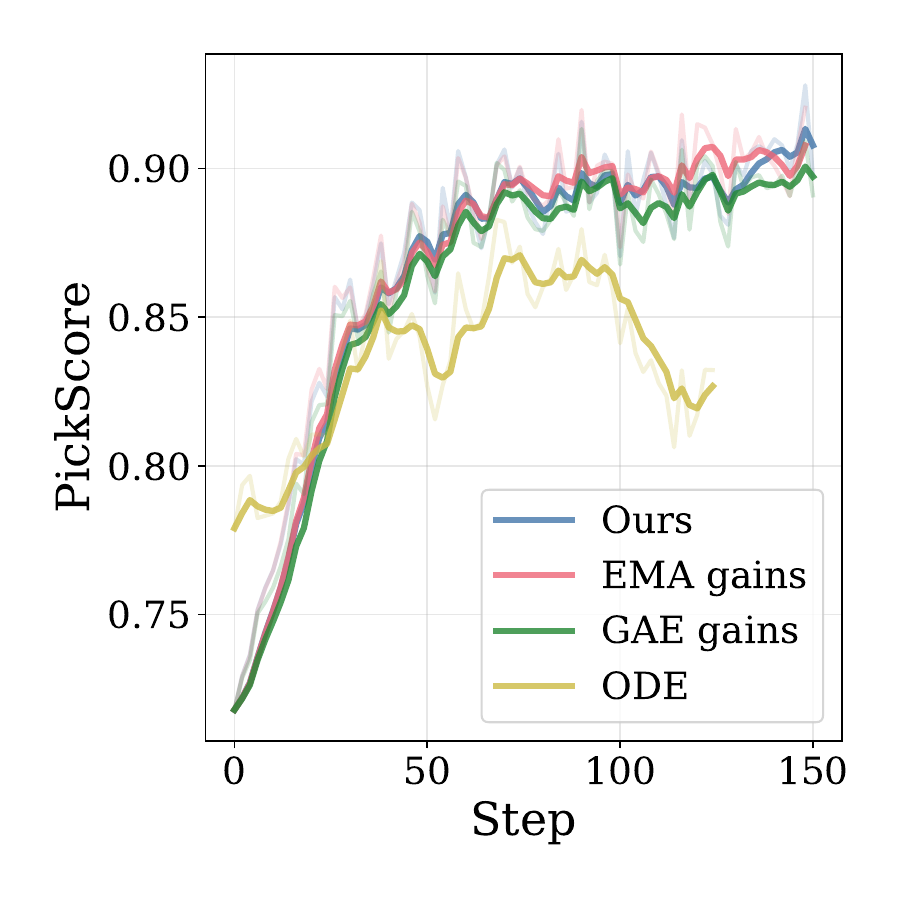}
    \caption{\textbf{Design variation comparison.} Reward vs. training iteration for different formulations of stepwise credit assignment on GenEval with PickScore reward. The standard gain formulation from the main paper matches all alternatives, demonstrating that preserving the natural temporal structure of diffusion gains is the most effective credit assignment.}
    \label{fig:design_variations}
\end{figure}

\cref{fig:design_variations} shows reward curves for all design variations compared to our standard gain formulation. No variation shows significant improvement over the standard formulation.

The superior performance of using advantage calculated directly on the gains suggests that the natural temporal structure of diffusion generation should be preserved rather than normalized away. EMA centering and GAE both don't significantly improve with this structure, while progressive distillation is far less stable than the SDE formulation. So, in the main paper we keep the original, simpler implementation.

\section{Improved SDE}
\label{app:improved_sde}

\begin{figure}[t]
\centering
\includegraphics[width=1.0\linewidth]{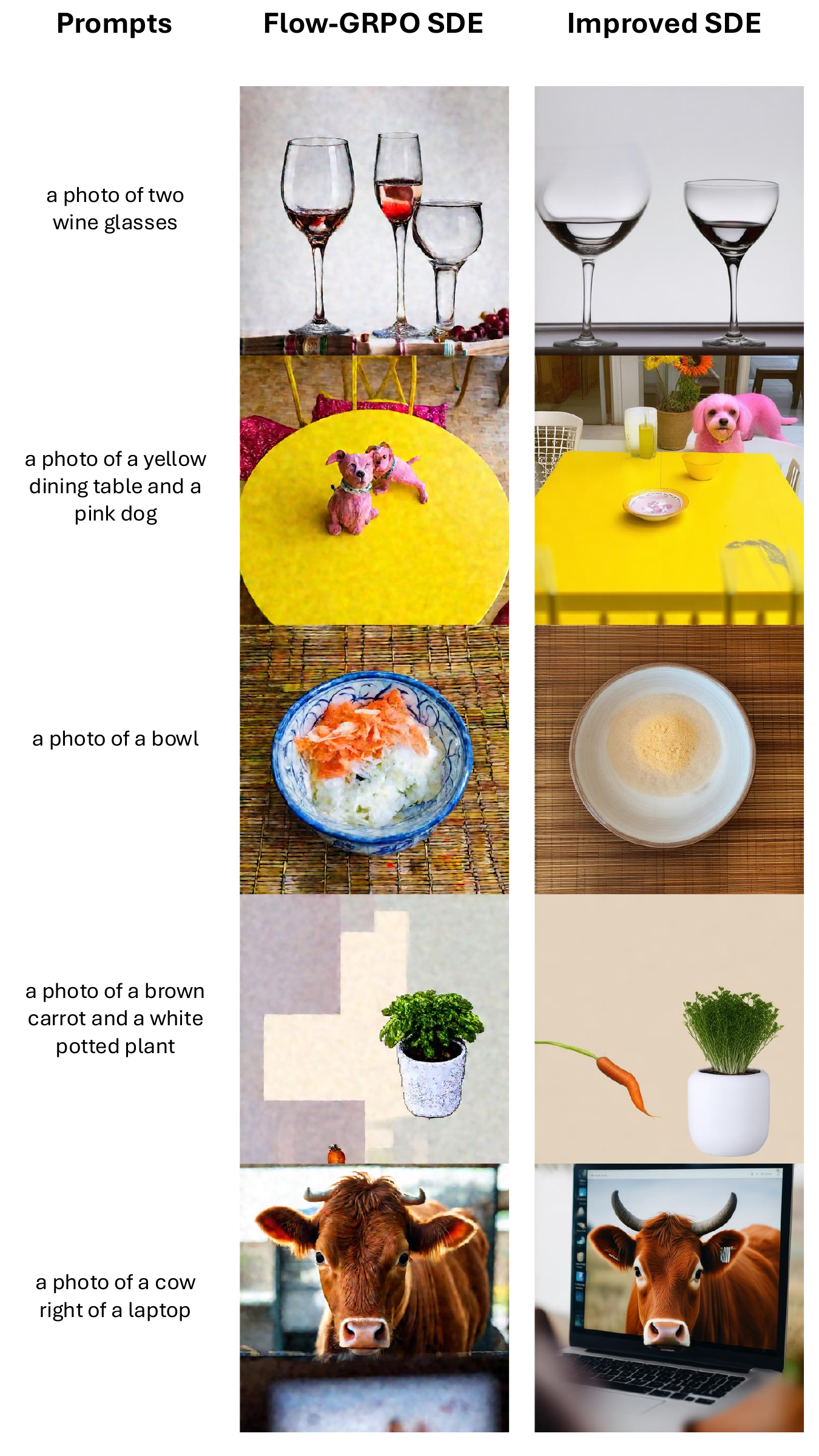}
\caption{\textbf{Improved SDE produces cleaner images.}
Qualitative comparison between Flow-GRPO SDE (middle) and our DDIM-inspired SDE (right) from \cref{subsec:improved_sde}.
The improved formulation generates images that are much less noisy while maintaining stochasticity for policy gradients.}
\label{fig:sde_comparison}
\end{figure}

Flow-GRPO's SDE generates noisy intermediate samples that degrade reward signal quality. While the SDE in~\cref{eq:sde} is theoretically sound for RL exploration, the noise injection mechanism produces visibly corrupted images at intermediate steps, see~\cref{fig:sde_comparison}. Since reward models are typically trained on clean images, this noise substantially reduces the informativeness of reward signals, slowing RL convergence. We address this by replacing Flow-GRPO's SDE with a DDIM-inspired alternative that provides exploration while producing cleaner samples.

We construct an SDE that interpolates between deterministic ODE sampling and stochastic exploration while preserving the variance structure of the flow. Drawing inspiration from DDIM~\citep{song2021denoising}, we define the transition from step $t$ to $t-\Delta t$ as $x_{t-\Delta t} = $
\begin{align}
   (1-(t-\Delta t))\,\hat x_0 + \sqrt{(t-\Delta t)^2 - \sigma_t^2}\,\hat x_1 + \sigma_t \epsilon
    \label{eq:ddim_sde}
\end{align}
where $\hat x_0 = x_t - t v_t$ and $\hat x_1 = x_t + (1-t)v_t$ are the predicted clean image and noise respectively given the flow prediction $v_t := v_\theta(x_t, t, c)$. Stochasticity is controlled with $\sigma_t$ and $\epsilon \sim \calN(0,I)$. When $\sigma_t = 0$, this recovers the deterministic ODE; when $\sigma_t > 0$, RL exploration is enabled through controlled noise injection.
\footnote{This is exactly equal to the DDIM update if we let $\alpha_{t-\Delta t} = (1-(t-\Delta t))^2$ and $\beta_t = (t-\Delta t)^2$, where $\alpha_t$ and $\beta_t$ are the signal and noise strengths in the paper respectively.}

Rewriting~\cref{eq:euler_maruyama} using $x_t = (1-t)\hat x_0 + t\hat x_1$, we obtain $x_{t-\Delta t} = $
\begin{align}
    (1-(t-\Delta t))\hat x_0 + \left((t-\Delta t) + \frac{\sigma_t^2 \Delta t}{2t}\right)\hat x_1 + \sigma_t \sqrt{\Delta t}\,\epsilon
    \label{eq:flow_rewritten}
\end{align}
Comparing with~\cref{eq:ddim_sde}, the noise coefficients differ by $\sqrt{(t-\Delta t)^2 - \sigma_t^2} \approx (t-\Delta t) - \frac{\sigma_t^2}{2(t-\Delta t)}$ via Taylor expansion. As $\Delta t \to 0$ and $\sigma_t \to 0$, the two formulations converge.

\subsection{Variance-Preserving Property}
The DDIM SDE exactly preserves the marginal variance at each step, while Flow-GRPO's SDE inflates it. For rectified flow, the training interpolation $x_t = (1-t)x_0 + tx_1$ implies that $\Var(x_t|x_0) = t^2$. Computing the variance of~\cref{eq:ddim_sde}:
\begin{align}
    &\Var(x_{t-\Delta t}|\hat x_0) \\
    &= ((t-\Delta t)^2 - \sigma_t^2)\Var(\hat x_1) + \sigma_t^2
    \label{eq:ddim_variance}
\end{align}
If $\var(\hat x_1)=1$, this exactly matches the expected variance. In contrast, from~\cref{eq:flow_rewritten}:
\begin{align}
    \Var_{\text{Flow}}(x_{t-\Delta t}|x_0) &= \left((t-\Delta t) + \frac{\sigma_t^2 \Delta t}{2t}\right)^2 + \sigma_t^2 \Delta t \nonumber\\
    &> (t-\Delta t)^2 \quad \text{for any } \sigma_t > 0, \Delta t > 0
    \label{eq:flow_variance}
\end{align}
This excess variance accumulates across steps, producing visibly noisier trajectories that confound reward evaluation.

\subsection{Adaptive Noise Schedule}
Setting $\sigma_t^2 = (t-\Delta t)^2$ in~\cref{eq:ddim_sde} yields:
\begin{align}
    x_{t-\Delta t} = (1-(t-\Delta t))\,\hat x_0 + (t-\Delta t)\epsilon
\end{align}
which completely discards the predicted noise $\hat x_1$. However, during training, the model learns that $x_t = (1-t)x_0 + tx_1$ where both $x_0$ and $x_1$ contain complementary information about the data distribution. At test time, both $\hat x_0$ and $\hat x_1$ carry entangled information about the true clean image, i.e.,~$\hat x_1$ is not merely noise to be replaced, but rather encodes structure that the model has learned to extract from $x_t$. Fully replacing $\hat x_1$ with independent noise $\epsilon$ produces samples of the form $(1-t)\hat x_0 + t\epsilon$, which lie outside the model's training distribution and yield poor velocity predictions $v_t$.

Setting $\sigma_t = \eta(t-\Delta t)$ retains a $(1-\eta^2)$ fraction of the predicted noise structure. This interpolates between deterministic sampling ($\eta=0$) and full replacement ($\eta=1$), providing RL exploration while maintaining samples within the model's learned distribution.

However, this uniform schedule treats all steps equally despite their differential impact on the final output. Noise injected at step $t$ propagates through the remaining denoising steps and affects $x_0$ through the accumulated flow. To quantify this sensitivity, consider the Jacobian $J_t = \partial x_0/\partial x_t$, which evolves according to the tangent flow equation:
\begin{align}
    \frac{dJ_t}{dt} = J_t \cdot \frac{\partial v_t(x_t,t,c)}{\partial x_t}, \quad J_0 = I
\end{align}
While exact computation is intractable, first-order analysis suggests $\|J_t\|_F^2 \approx 1 + ct$ for some constant $c > 0$ when $\|\partial v_t/\partial x_t\|_F = O(1)$ (typical for normalized neural networks). This indicates that perturbations at earlier steps (larger $t$) have amplified influence on the final image.

Setting $\sigma_t = \eta t\sqrt{1-t}$ provides this compensation: the $\sqrt{1-t}$ factor naturally downweights exploration at early steps (when $t \to 1$, $\sqrt{1-t} \to 0$) where sensitivity is highest, while allowing more exploration at later steps where perturbations have localized effects. This makes the sensitivity-weighted exploration $\sigma_t^2 \cdot (1+ct)$ more uniform across the trajectory, ensuring that all steps contribute roughly equally to the RL exploration budget.

\subsection{RL Objective}
We apply the same GRPO objective as~\citet{liu2025flow} but with the DDIM SDE as our policy. The marginal probability for~\cref{eq:ddim_sde} is:
\begin{align}
    \pi_\theta(x_{t-\Delta t} | x_t, c) = \calN\left(x_{t-\Delta t}; \mu_t, \sigma_t^2 I\right)
    \label{eq:ddim_marginal}
\end{align}
where $\mu_t = (1-(t-\Delta t))\hat x_0 + \sqrt{(t-\Delta t)^2 - \sigma_t^2}\,\hat x_1$. We optimize the objective in~\cref{eq:flow-grpo} using this policy, inheriting the same clipping, KL regularization, and group-relative advantages from Flow-GRPO. The cleaner intermediate samples from our SDE enable more accurate reward evaluation, particularly for vision-based reward models sensitive to image quality.

\section{Ablation Studies}
\label{app:ablations}

We conduct ablation studies to validate key design choices in Stepwise-Flow-GRPO: the gain normalization strategy and the number of ODE substeps for computing intermediate clean image estimates. All experiments use PickScore reward on the GenEval dataset with the same training protocol as our main experiments.

\subsection{Gain Normalization Strategy}

A critical design choice is whether to normalize gains $g_t^i$ globally across all steps and trajectories (joint normalization) or separately at each step (per-step normalization). This decision affects how the temporal magnitude of gains is preserved in the final advantages.

\textbf{Joint normalization (our method)}: Compute mean and standard deviation across all steps and trajectories:
    \begin{align}
        \tilde{A}_t^i &= \frac{g_t^i - \mu_t}{\sigma_{\text{global}}}, \\
        \mu_{\text{global}}
        &= \frac{1}{NT}\sum_{j,k} g_k^j,\quad \sigma_{\text{global}} = \frac{1}{NT}\sum_{j,k} (g_k^j - \mu_{\text{global}})^2
    \end{align}
    This preserves the relative magnitudes across steps, so early gains with naturally larger values receive proportionally larger advantages.

\textbf{Stepwise normalization}: Compute mean and standard deviation separately for each step:
    \begin{align}
        \tilde{A}_t^i &= \frac{g_t^i - \mu_t}{\sigma_t}, \\
        \mu_t &= \frac{1}{N}\sum_j g_t^j,\quad \sigma_t = \frac{1}{N}\sum_{j} (g_t^j - \mu_t)^2
    \end{align}
    This equalizes the importance of all steps regardless of their natural gain magnitudes.

\begin{figure}[t]
    \centering
    \includegraphics[width=\linewidth]{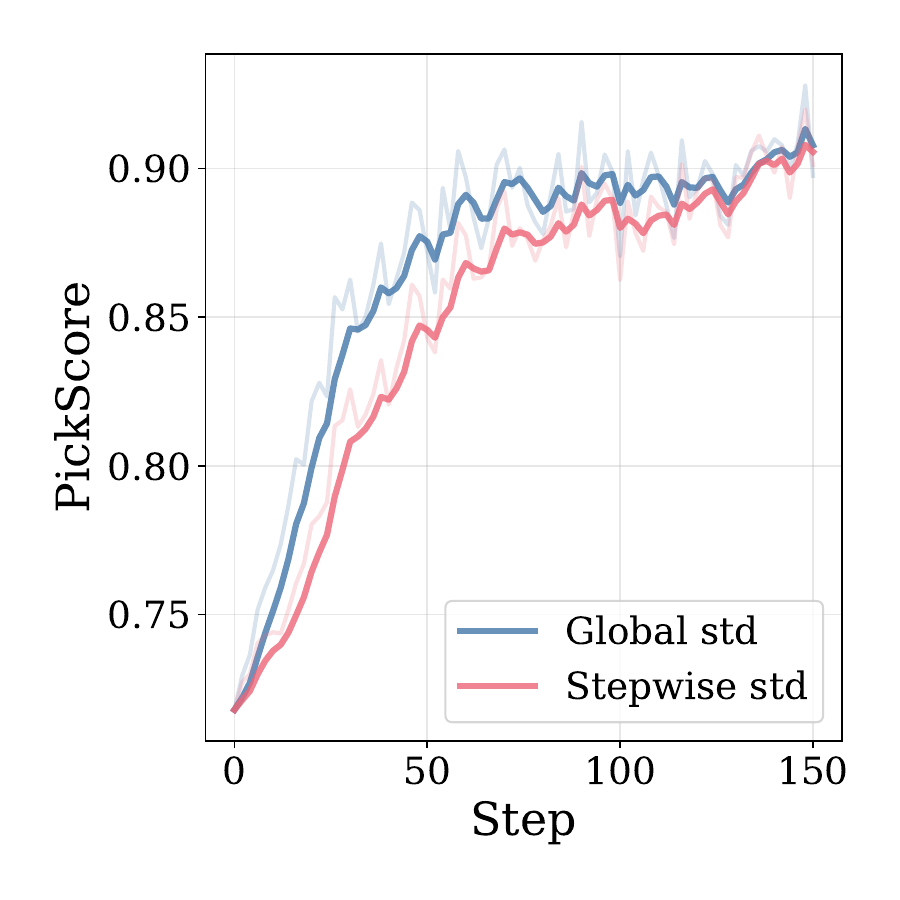}
    \caption{\textbf{Joint normalization preserves temporal structure and accelerates convergence.} Reward vs. training iteration comparing joint normalization (global mean/std across all steps and trajectories) against per-step normalization (separate mean/std for each step).}
    \label{fig:normalization_ablation}
\end{figure}

\cref{fig:normalization_ablation} shows that joint normalization significantly outperforms per-step normalization in convergence speed. Both methods eventually converge to similar final rewards, indicating that stepwise normalization doesn't improve final quality, only slows down learning.

\subsection{Number of Denoising Substeps}

Computing intermediate reward estimates $r_t^i = R(\hat{x}_0(t), c)$ requires denoising from noisy state $x_t$ to obtain a clean image estimate $\hat{x}_0(t)$. The number of ODE substeps $T'$ controls the tradeoff between estimate quality and computational cost.

\begin{figure}[t]
    \centering
    \includegraphics[width=0.49\linewidth]{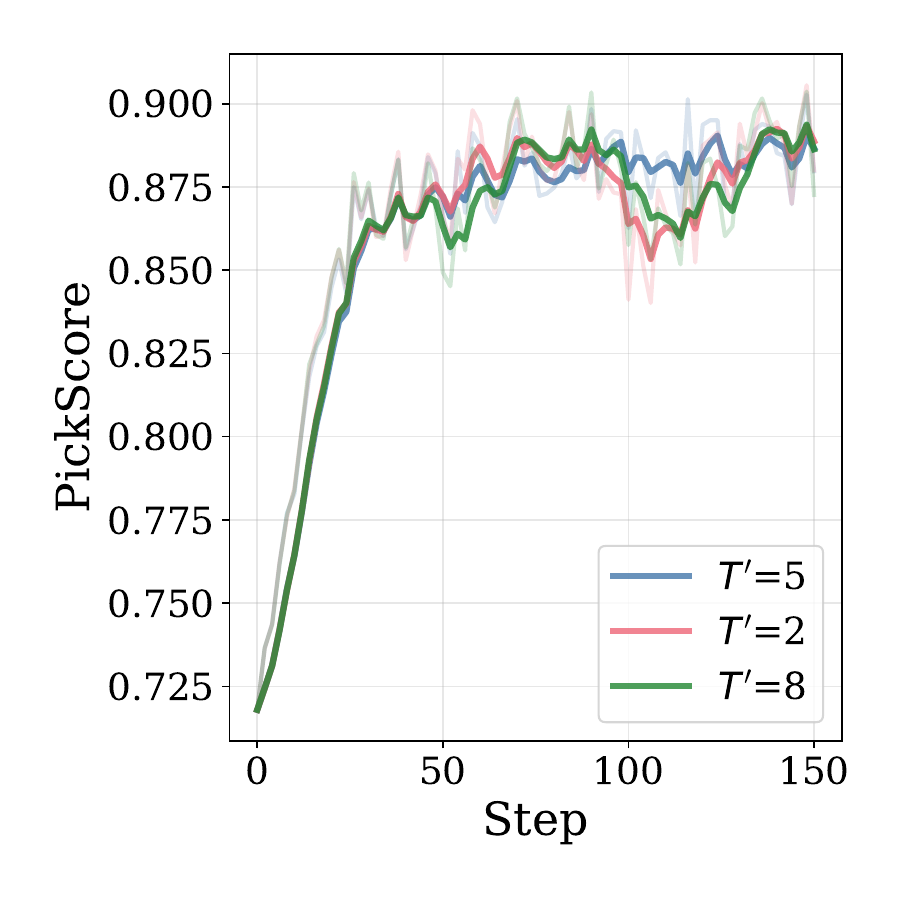}
    \includegraphics[width=0.49\linewidth]{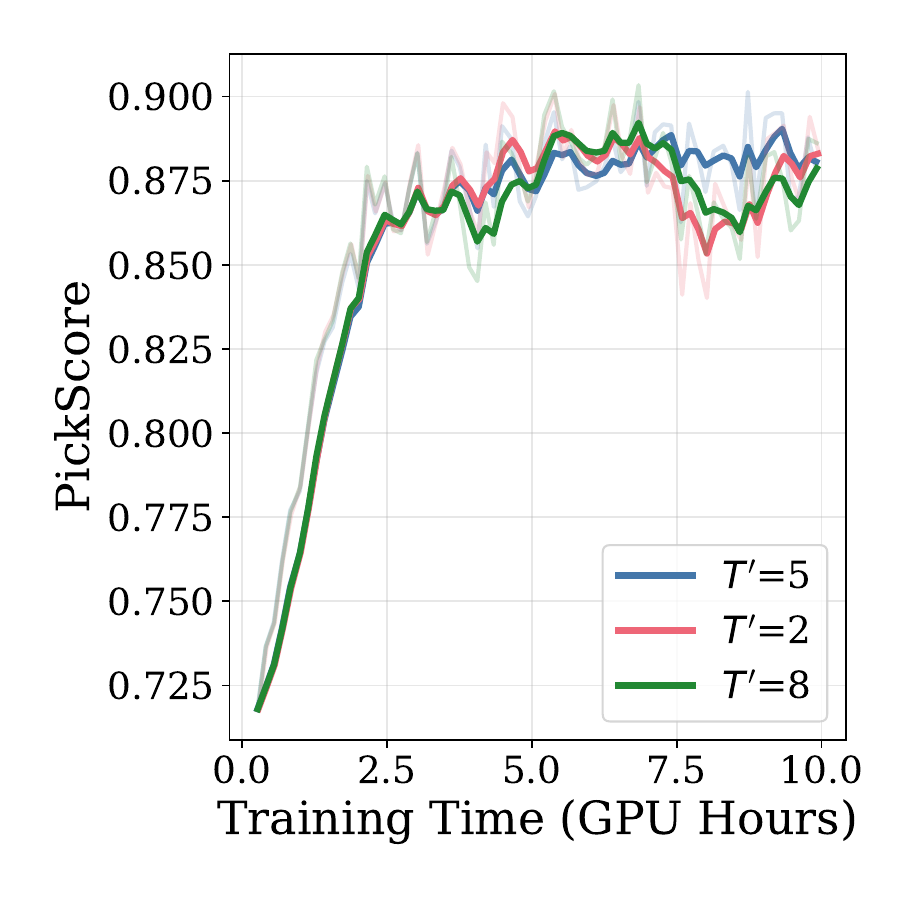}
    \caption{\textbf{Stepwise-Flow-GRPO is robust to the number of denoising substeps.} Reward vs. training iteration (top) and GPU Hours (bottom) for different numbers of substeps $T' \in \{2, 5, 8\}$ used to compute $\hat{x}_0(t)$. All settings achieve similar convergence speed and final performance, validating that our method is not sensitive to this hyperparameter choice.}
    \label{fig:substep_ablation}
\end{figure}

We compare $T' \in \{2, 5, 8\}$ on PickScore reward using the GenEval dataset, measuring both reward convergence and wall-clock time. \cref{fig:substep_ablation} shows that all choices of $T'$ achieve similar performance in both training iterations and wall-clock time, demonstrating that our method is robust to this hyperparameter. We select $T'=5$ as our default, though the results suggest practitioners can adjust this based on their computational constraints without significantly impacting final performance.

\section{Runtime Analysis}
\label{app:timing}

Stepwise-Flow-GRPO incurs additional computational cost from: (1) computing intermediate estimates $\hat{x}_0(t)$ via $T'=5$ ODE substeps from each $x_t$, and (2) evaluating the reward model on these estimates at each step. \cref{tab:timing} shows per-iteration timing on 8 NVIDIA A100 GPUs with batch size 16.

\textbf{Generation overhead.}
Our method requires approximately 1.8-2.4$\times$ more generation time than Flow-GRPO due to additional ODE integration (50 extra steps for $T=10$ steps). These intermediate denoising steps are embarrassingly parallel and batch efficiently.

\textbf{Reward evaluation overhead.}
The overhead varies by reward model: PickScore (lightweight CNN) adds minimal cost (2.8s vs 1.4s), ImageReward shows ~10$\times$ overhead (8.2s vs 0.7s), and UnifiedReward (7B VLM) dominates computation (114.6s vs 48.8s). For UnifiedReward, we use 8 separate A100 80GB GPUs with SGLang for efficient batched inference.

\begin{table}[t]
\centering
\small
\begin{tabular}{l|cc}
\toprule
\textbf{Method (Reward)} & \textbf{Generation (s)} & \textbf{Reward eval (s)} \\
\midrule
Ours (PickScore) & 24.4 $\pm$ 0.0 & 2.8 $\pm$ 0.0 \\
Flow-GRPO (PickScore) & 13.7 $\pm$ 0.9 & 1.4 $\pm$ 1.3 \\
Ours (ImageReward) & 29.8 $\pm$ 0.2 & 8.2 $\pm$ 0.2 \\
Flow-GRPO (ImageReward) & 7.9 $\pm$ 0.0 & 0.7 $\pm$ 0.0 \\
Ours (UnifiedReward) & 136.1 $\pm$ 5.8 & 114.6 $\pm$ 5.8 \\
Flow-GRPO (UnifiedReward) & 56.0 $\pm$ 30.9 & 48.8 $\pm$ 30.9 \\
\bottomrule
\end{tabular}
\caption{\textbf{Per-iteration timing breakdown} in seconds per training iteration, averaged over multiple runs.}
\label{tab:timing}
\end{table}

\textbf{Implementation optimizations.}
We implement several optimizations to minimize overhead:
(1) Batched intermediate denoising: All $T'=5$ substeps for a given $x_t$ are batched together.
(2) Parallel reward evaluation: Intermediate estimates $\{\hat{x}_0(t)\}_{t=1}^T$ are evaluated in parallel across steps.
(3) Asynchronous execution: Generation and reward evaluation are pipelined when possible.

Crucially, despite the 1.8-2.4$\times$ per-iteration slowdown, Stepwise-Flow-GRPO achieves \emph{faster overall convergence in wall-clock time} across all settings (\cref{fig:wallclock} in main paper). The superior sample efficiency—requiring 2-3$\times$ fewer training iterations to reach target rewards—more than compensates for the per-iteration overhead. In practice, our method converges 20-40\% faster in total wall-clock time depending on the reward function, demonstrating that the improved learning signal justifies the additional computation.

\section{Additional Results}
\label{app:additional_results}

We provide extended experimental results including: (1) OCR text rendering evaluation, (2) longer training runs with the GenEval reward, and (3) extended training with UnifiedReward. These experiments validate that our method's advantages extend across diverse reward functions and training durations.

\subsection{OCR Text Rendering}

We evaluate on a specialized OCR dataset using a combined reward: 80\% OCR accuracy + 20\% PickScore. This challenging compositional task requires rendering readable text while maintaining visual quality.

\begin{figure}[t]
    \centering
    \includegraphics[width=0.49\linewidth]{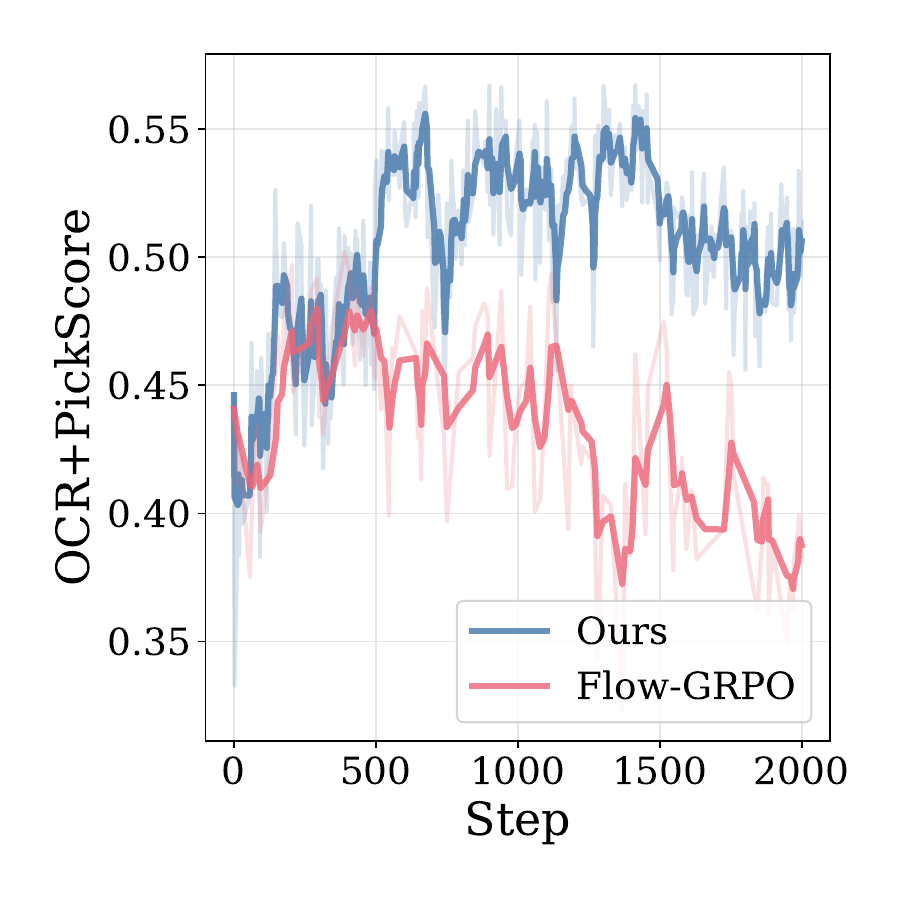}
    \includegraphics[width=0.49\linewidth]{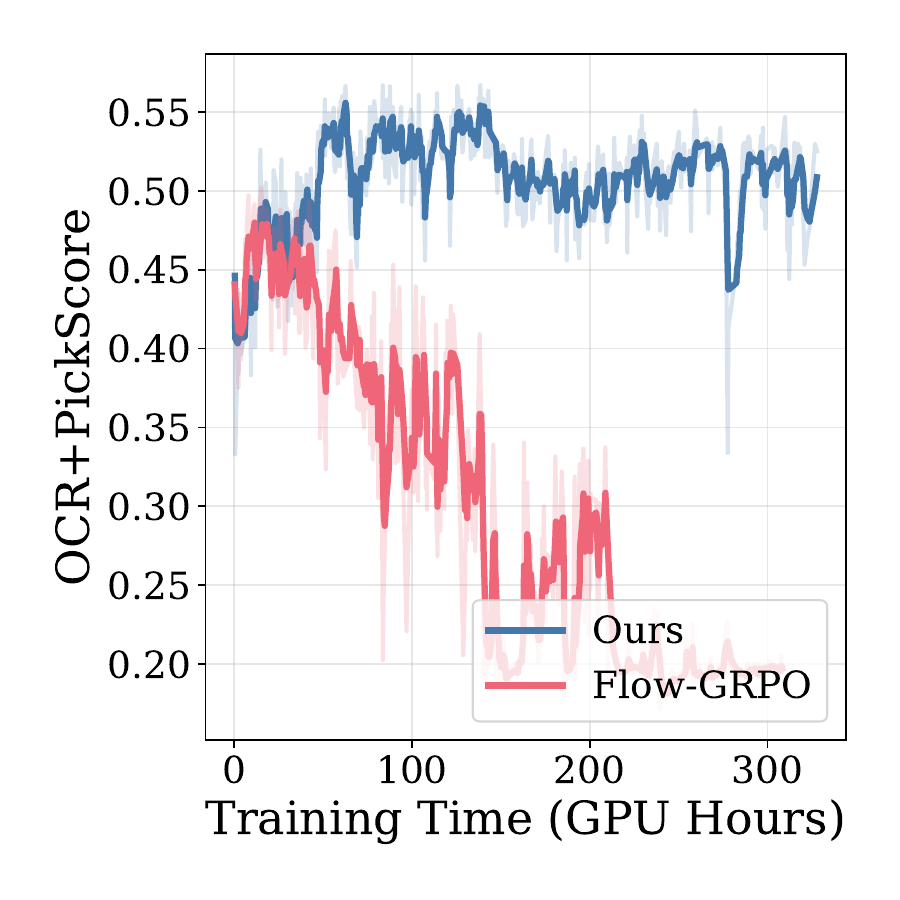}
    \caption{\textbf{OCR text rendering results.} Reward vs. training iteration (top) and wall-clock time (bottom) using combined OCR+PickScore reward (80\% OCR, 20\% PickScore). Flow-GRPO diverges after ~500 steps while Stepwise-Flow-GRPO continues improving, demonstrating superior stability and final performance on compositional text rendering.}
    \label{fig:ocr_results}
\end{figure}

\cref{fig:ocr_results} shows that Stepwise-Flow-GRPO substantially outperforms Flow-GRPO. Flow-GRPO diverges after $\sim$500 steps, while our method continues improving and plateaus at a significantly higher reward. This demonstrates particularly strong benefits for hierarchical compositional tasks where early steps establish structure (letter shapes, spacing) that later steps refine.

\subsection{Extended GenEval Training}

\begin{figure}[t!]
    \includegraphics[width=\linewidth]{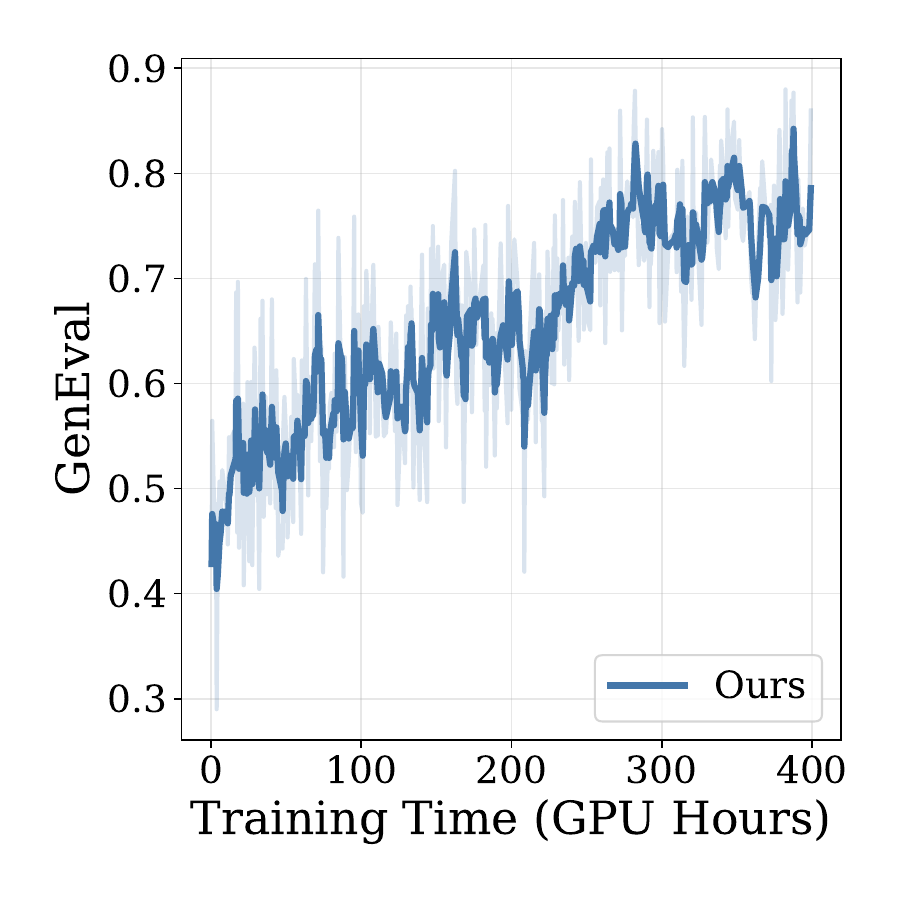}
    \caption{\textbf{Extended GenEval training.} GenEval overall score vs. wall-clock time for 400 GPU hour runs. Stepwise-Flow-GRPO achieves 0.87, substantially outperforming Flow-GRPO (0.72) and approaching state-of-the-art autoregressive models. The widening performance gap demonstrates that stepwise credit assignment provides increasing benefits at high performance levels.}
    \label{fig:geneval_long}
\end{figure}

To test scalability, we conduct a 400 GPU hour training run on GenEval. \cref{fig:geneval_long} shows that after 400 hours, Stepwise-Flow-GRPO achieves 0.87 overall GenEval score—substantially outperforming Flow-GRPO (0.72 from paper) and even beats state-of-the-art models like GPT-4o (0.84).

The performance gap \emph{widens} with extended training, particularly on challenging categories: our method achieves 0.89 on counting, 0.73 on spatial positioning, and 0.80 on attribute binding. These categories require precise compositional decisions that benefit most from accurate credit assignment. This suggests that appropriate credit assignment becomes more critical as models approach high performance levels.

\subsection{UnifiedReward Training}

We validate sustained advantages with UnifiedReward-7b-v1.5, a large vision-language model that evaluates caption alignment and visual quality.

UnifiedReward prompt: \textit{"You are given a text caption and a generated image based on that caption. Your task is to evaluate this image based on two key criteria: (1) Alignment with the Caption: Assess how well this image aligns with the provided caption. Consider the accuracy of depicted objects, their relationships, and attributes. (2) Overall Image Quality: Examine the visual quality including clarity, detail preservation, color accuracy, and aesthetic appeal. Assign a score from 1 to 5 after 'Final Score:'."}

\begin{figure}[t!]
    \centering
    \includegraphics[width=\linewidth]{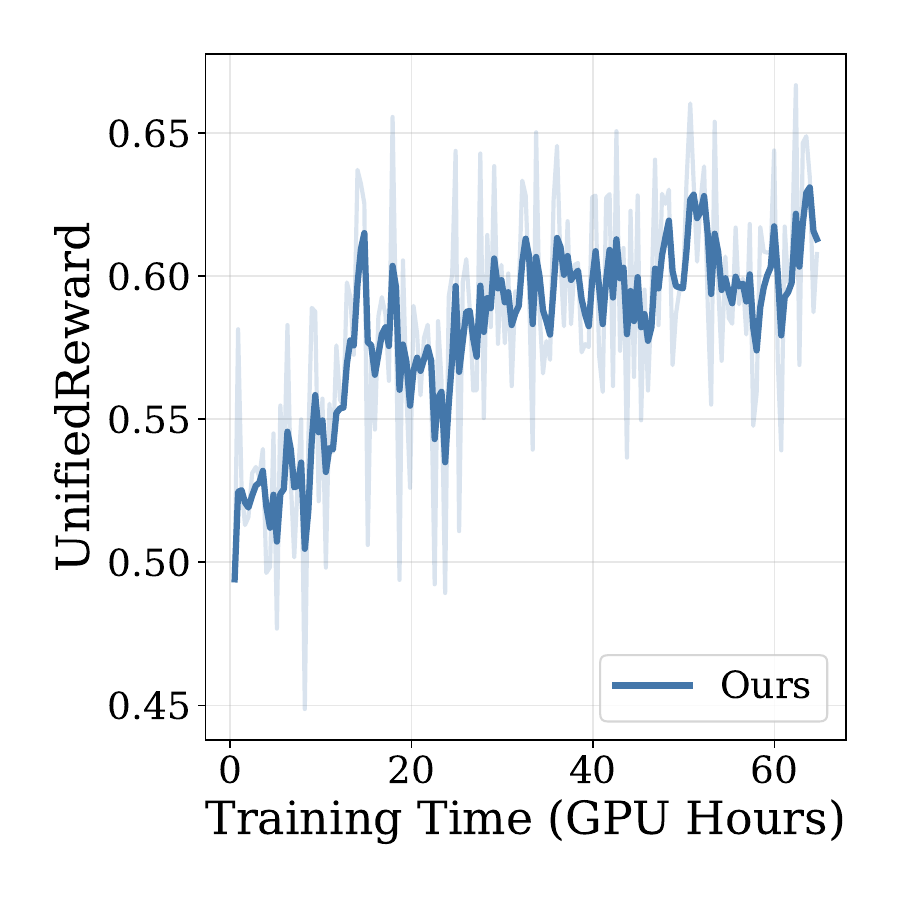}
    \caption{\textbf{UnifiedReward training.} Reward vs. wall-clock time for 60 GPU hour run on GenEval using UnifiedReward-7b-v1.5. Stepwise-Flow-GRPO trains stably while Flow-GRPO diverges with this reward function, demonstrating superior robustness with complex VLM-based rewards.}
    \label{fig:unifiedreward_long}
\end{figure}

\cref{fig:unifiedreward_long} shows that after 60 GPU hours, Stepwise-Flow-GRPO achieves 0.74 GenEval score (\cref{tab:geneval_extended}) with smooth, stable training curves. Our method maintains consistent efficiency advantages throughout extended training.

\textbf{Training stability.}
Notably, Flow-GRPO with the standard SDE formulation consistently diverged when training with UnifiedReward, preventing stable optimization. In contrast, our method trains stably throughout, demonstrating that stepwise credit assignment provides not only efficiency improvements but also fundamental stability benefits when using complex reward models. This increased robustness is particularly valuable for large VLM-based rewards where gradient noise can be substantial.

\begin{table*}[t]
\small
\centering
\begin{tabular}{l | c | c c c c c c}
\toprule
\textbf{Model} & \textbf{Overall} & \textbf{Single Obj.} & \textbf{Two Objs.} & \textbf{Counting} & \textbf{Colors} & \textbf{Position} & \textbf{Attr. Binding} \\
\midrule
\multicolumn{8}{c}{\textit{Pretrained Models}} \\
\midrule
SD3.5-M (cfg=1.0) & 0.28 & 0.71 & 0.23 & 0.15 & 0.45 & 0.05 & 0.08 \\
SD3.5-M (cfg=4.5) & 0.63 & 0.98 & 0.78 & 0.50 & 0.81 & 0.24 & 0.52 \\
\midrule
\multicolumn{8}{c}{\textit{Standard Training Duration}} \\
\midrule
Flow-GRPO (cfg=1.0, PickScore) & 0.60 & 0.96 & 0.73 & 0.67 & 0.67 & 0.21 & 0.35 \\
\textbf{Ours (cfg=1.0, PickScore)} & 0.60 & 0.96 & 0.75 & 0.67 & 0.67 & 0.21 & 0.34 \\
Flow-GRPO (cfg=4.5, PickScore) & 0.68 & 0.98 & 0.82 & 0.64 & 0.82 & 0.24 & 0.59 \\
\textbf{Ours (cfg=4.5, PickScore)} & \textbf{0.71} & \textbf{0.98} & \textbf{0.85} & \textbf{0.70} & \textbf{0.82} & \textbf{0.29} & \textbf{0.59} \\
\midrule
\multicolumn{8}{c}{\textit{Extended Training}} \\
\midrule
Flow-GRPO (cfg=4.5, GenEval, 400 GPU hrs) & 0.72 & -- & -- & -- & -- & -- & -- \\
\textbf{Ours (cfg=4.5, UnifiedReward, 60 GPU hrs)} & 0.74 & 0.99 & 0.89 & 0.73 & 0.83 & 0.34 & 0.66 \\
\textbf{Ours (cfg=4.5, GenEval, 400 GPU hrs)} & \textbf{0.87} & \textbf{0.99} & \textbf{0.93} & \textbf{0.89} & \textbf{0.87} & \textbf{0.73} & \textbf{0.80} \\
\midrule
\multicolumn{8}{c}{\textit{Reference: State-of-the-art Models}} \\
\midrule
Janus-Pro-7B & 0.80 & 0.99 & 0.89 & 0.59 & 0.90 & 0.79 & 0.66 \\
SANA-1.5 4.8B & 0.81 & 0.99 & 0.93 & 0.86 & 0.84 & 0.59 & 0.65 \\
GPT-4o & 0.84 & 0.99 & 0.92 & 0.85 & 0.92 & 0.75 & 0.61 \\
\bottomrule
\end{tabular}
\caption{\textbf{Complete GenEval results.} After 400 GPU hours, our method achieves 0.87 overall, substantially outperforming Flow-GRPO (0.72) and approaching GPT-4o (0.84). Flow-GRPO extended results from \cite{liu2025flow}; reference models from respective papers.}
\label{tab:geneval_extended}
\end{table*}

\begin{figure*}[t]
\centering
\includegraphics[width=1.0\linewidth]{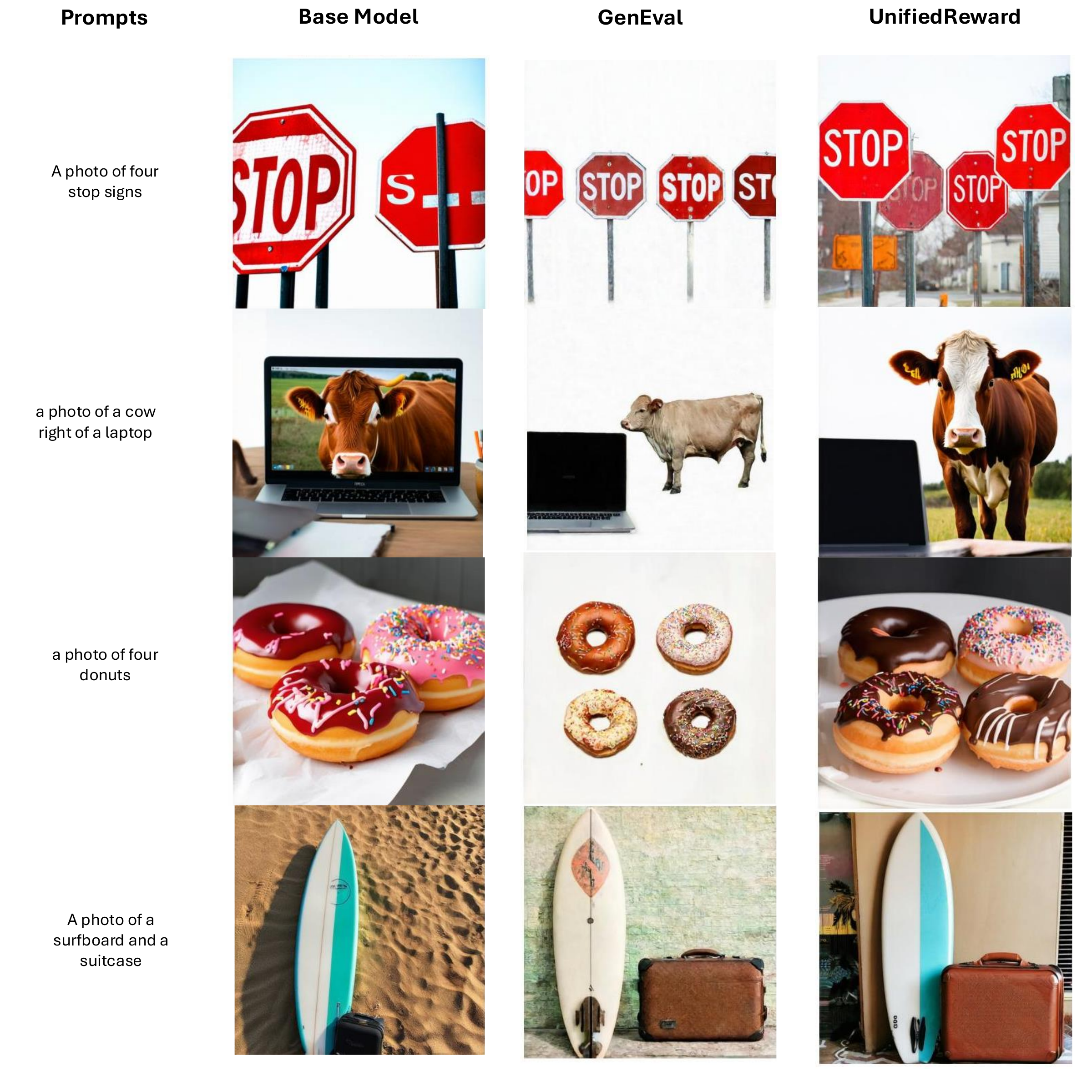}
\caption{\textbf{Qualitative comparison across training objectives.} Generated images from GenEval prompts using base SD3.5-M (left), GenEval reward training (middle), and UnifiedReward training (right). While GenEval reward training improves prompt adherence and benchmark scores, UnifiedReward training produces higher overall visual quality and more photorealistic images.}
\label{fig:more_images}
\end{figure*}

\subsection{Qualitative Results}

\begin{figure*}[t]
\centering
\includegraphics[page=1, height=0.9\textheight]{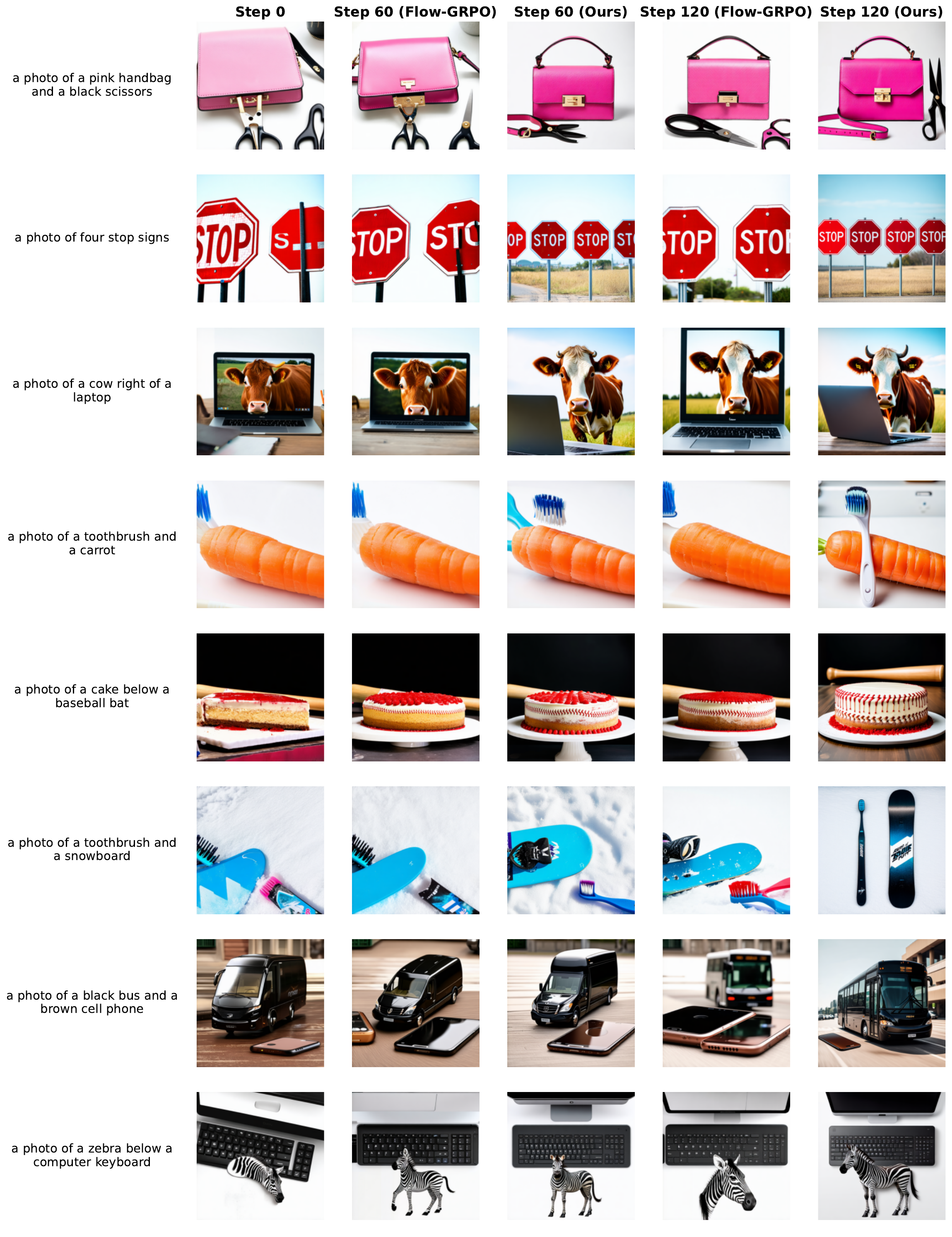}
\caption{\textbf{Extended qualitative comparison (page 1).} Each row shows a single prompt with the base model generation (step 0), followed by Flow-GRPO and Stepwise-Flow-GRPO (Ours) at training steps 60 and 120, both trained with PickScore reward. The gap is most visible at step 60, where our method demonstrates better counting, more plausible compositions, and higher image quality.}
\label{fig:qual_page1}
\end{figure*}

\begin{figure*}[t]
\centering
\includegraphics[page=2, height=0.9\textheight]{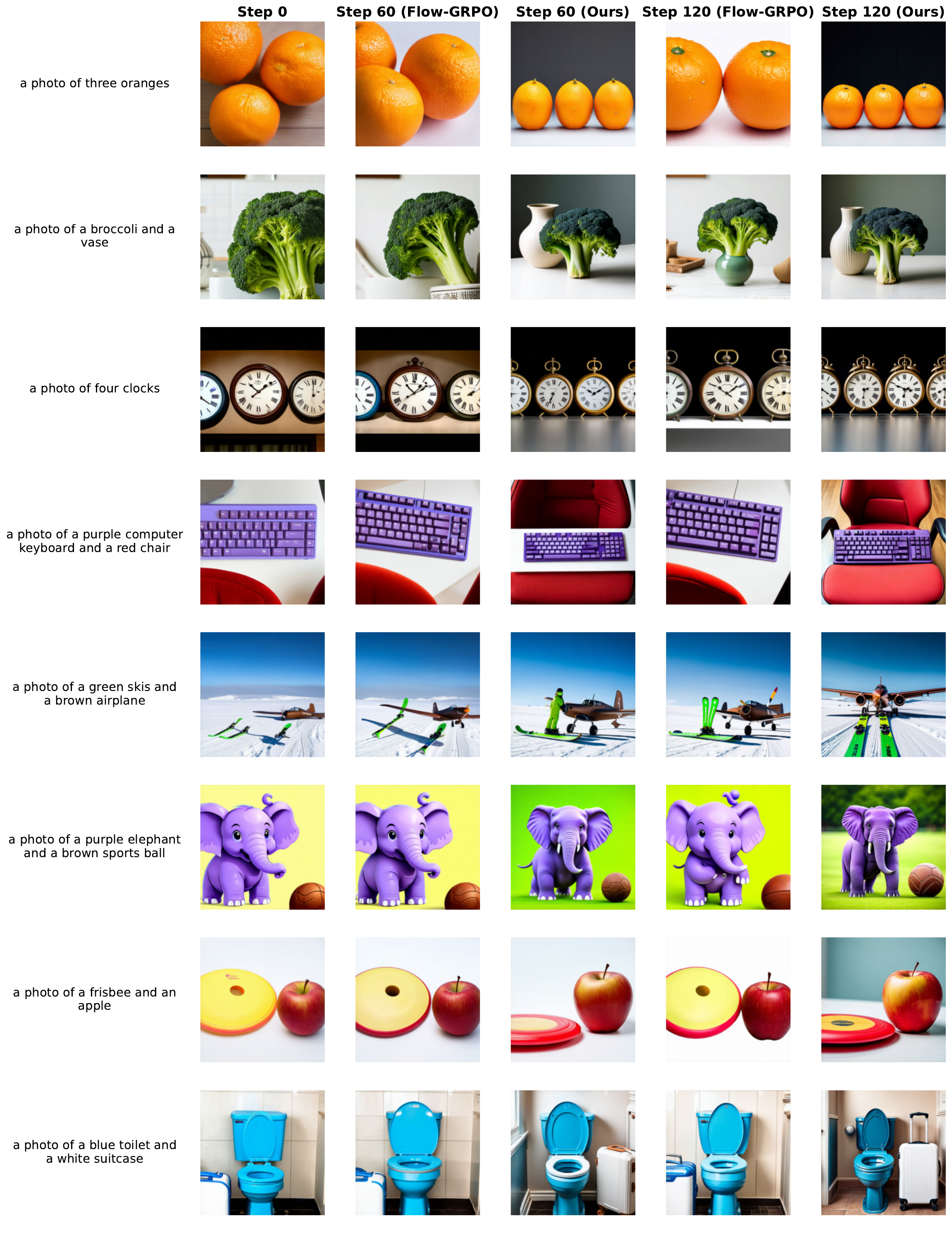}
\caption{\textbf{Extended qualitative comparison (page 2).} Same layout as \cref{fig:qual_page1}. Stepwise-Flow-GRPO consistently produces more accurate object counts, better spatial arrangements, and higher overall quality than Flow-GRPO across training.}
\label{fig:qual_page2}
\end{figure*}

\begin{figure*}[t]
\centering
\includegraphics[page=3, width=\linewidth]{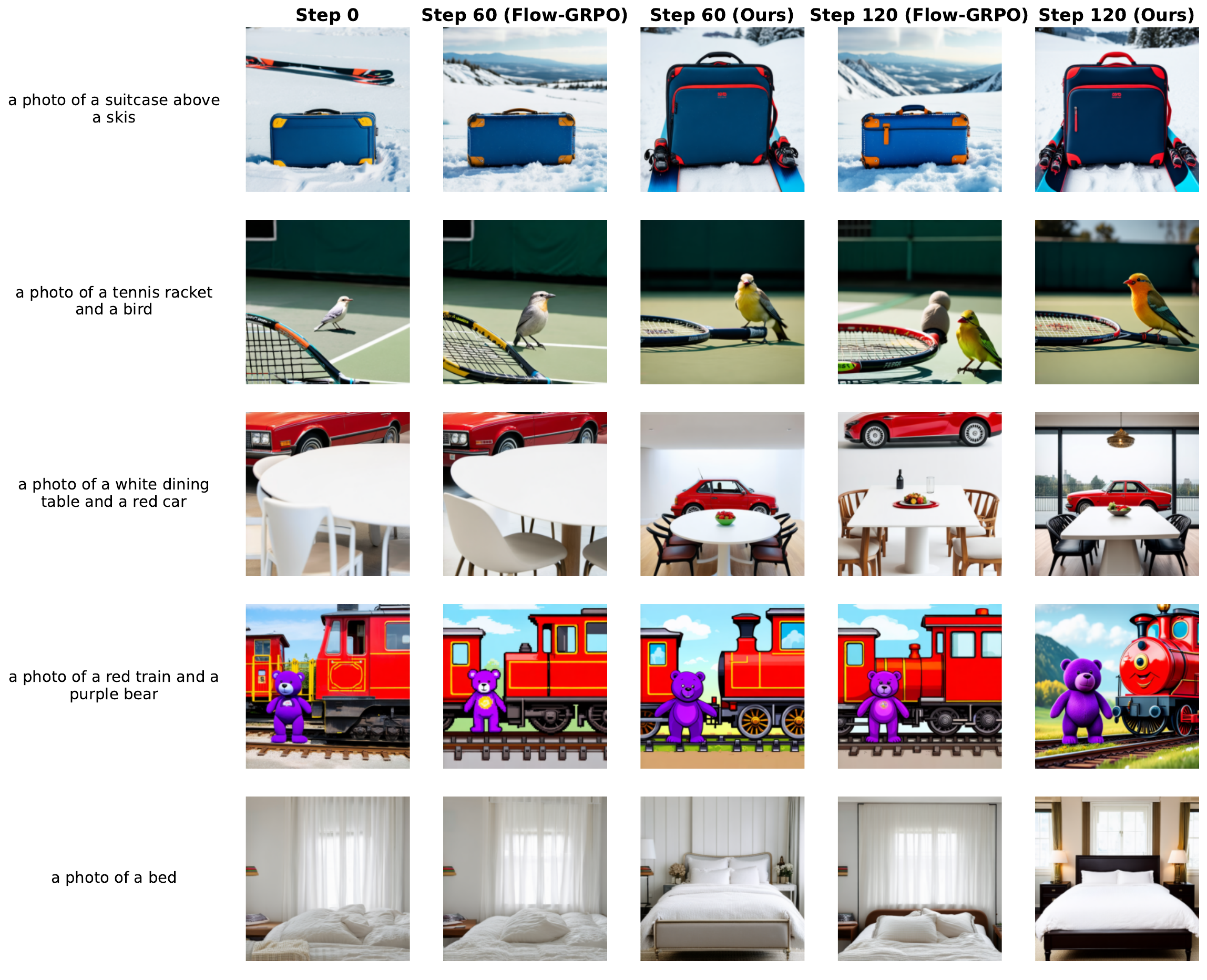}
\caption{\textbf{Extended qualitative comparison (page 3).} Same layout as \cref{fig:qual_page1}. Our method maintains qualitative advantages throughout training, with the largest visible differences at step 60.}
\label{fig:qual_page3}
\end{figure*}

\cref{fig:qual_page1,fig:qual_page2,fig:qual_page3} provide extended qualitative comparisons between Flow-GRPO and Stepwise-Flow-GRPO. Each row shows a single prompt with the base model output (step 0) and results from both methods at training steps 60 and 120. Our method produces consistently better images, with the most pronounced differences at step 60 when the performance gap is largest.

Stepwise-Flow-GRPO shows clear improvements in \textbf{counting} (e.g., ``three oranges'', ``four clocks'', ``a tennis racket and a bird''), \textbf{real-world dynamics} (e.g., ``broccoli and a vase'', ``a white dining table and a red car''), and \textbf{overall image quality} (e.g., ``a red train and a purple bear'', ``bed''). These improvements are consistent with our method's ability to assign credit to the denoising steps that establish composition and object placement, rather than uniformly rewarding the entire trajectory.

\section{Comparison with Concurrent Work}
\label{app:concurrent_work}

TempFlow-GRPO~\cite{he2025tempflow} and Granular-GRPO~\cite{zhou2025fine} are concurrent works that also address the uniform credit assignment limitation in Flow-GRPO. All three methods recognize that early denoising steps have outsized impact on final image quality, but the approaches differ in several key respects.

\noindent \textbf{What is optimized.}
TempFlow-GRPO and Granular-GRPO both use the standard GRPO advantage $\hat{A}^i = (R(x_0^i, c) - \text{mean})/\text{std}$, i.e., the normalized \emph{final reward}, with per-step attribution via trajectory branching. In contrast, we optimize telescoping gains $g_t^i = r_{t-1}^i - r_t^i$, directly rewarding each step's \emph{marginal improvement}. This captures causal contribution rather than terminal correlation.

\noindent \textbf{How early steps are emphasized.}
TempFlow-GRPO applies a hand-designed noise-level weighting $\text{Norm}(\sigma_t\sqrt{\Delta t})$ to discount advantages at later steps. Granular-GRPO optimizes only the first 8 of 16 steps, leaving later steps unoptimized. Our gains are \emph{data-dependent}: as shown in~\cref{fig:gains}, gain magnitudes naturally decrease as $t \to 0$, automatically concentrating optimization on early steps without manual schedules or step selection.

\noindent \textbf{Sampling efficiency.}
Both TempFlow-GRPO and Granular-GRPO use ODE$\to$SDE$\to$ODE branching, requiring $O(T^2 K)$ forward passes for $T$ steps and $K$ branches. We use the SDE throughout with few-step Tweedie estimation, which is embarrassingly parallel and empirically faster---running TempFlow-GRPO's released code at equivalent batch sizes, we observed our method to be approximately $1.5\times$ faster per epoch.

\end{document}